\documentclass{article}   
\usepackage{osajnl2} 

\usepackage{times}
\usepackage{psfig}
\usepackage{mysubfigure}
\usepackage{algorithm}
\usepackage{algorithmic}
\usepackage{epsfig}
\usepackage{graphicx}
\usepackage{amsmath}
\usepackage{amssymb}
\usepackage{multirow}
\usepackage{rotating}
\usepackage{mdwlist}

\newcommand{\BUT}{\vspace*{-0.25in}}

\newcommand{\mbn}{\mbox{\boldmath$ n\:$}}
\newcommand{\mbnS}{\mbox{\boldmath$ n\!_S$}}

\newcommand{\mba}{\mbox{\boldmath$ a\:$}}

\newcommand{\mbb}{\mbox{\boldmath$ b\:$}}

\newcommand{\mbrho}{\mbox{\boldmath$ \rho\:$}}

\newcommand{\mbe}{\mbox{\boldmath$ e\:$}}

\newcommand{\mbu}{\mbox{\boldmath$ u\:$}}
\newcommand{\mbv}{\mbox{\boldmath$ v\:$}}
\newcommand{\mbw}{\mbox{\boldmath$ w\:$}}
\newcommand{\mbr}{\mbox{\boldmath$ r\:$}}
\newcommand{\mbchi}{\mbox{\boldmath$ \chi\:$}}
\newcommand{\mbxi}{\mbox{\boldmath$ \xi\:$}}
\newcommand{\mbpsi}{\mbox{\boldmath$ \psi\:$}}
\newcommand{\mbepsilon}{\mbox{\boldmath$ \epsilon\:$}}
\newcommand{\mbeta}{\mbox{\boldmath$ \eta\:$}}
\newcommand{\mbzeta}{\mbox{\boldmath$ \zeta\:$}}

\newcommand{\mbx}{\mbox{\boldmath$ x\:$}}

\newcommand{\mbR}{\mbox{\boldmath$ R\:$}}
\newcommand{\mbM}{\mbox{\boldmath$ M\:$}}

\begin{document}
\title{Camera Calibration for Daylight Specular-Point Locus} 

\author{Mark S.\ Drew$^1$, Hamid Reza Vaezi Joze$^1$, and Graham D.\ Finlayson$^2$}
\address{$^1$School of Computing Science, Simon Fraser University,\\
   Vancouver, B.C., Canada V5A~1S6\\
         $^2$School of Computing Sciences, The University of East Anglia,\\
   Norwich, U.K., NR4~7TJ\\
       }

\maketitle

\vspace*{0.25in}
\begin{center}
	\bf April 2012
\end{center}

\vspace*{-0.25in}

\begin{abstract}
In this paper we present a new camera calibration method aimed at finding a
straight-line
locus, in a special colour feature space, that is traversed by daylights and as
well also approximately
followed by specular points.
The aim of the calibration is to enable recovering the colour of the illuminant
in a scene, using the calibrated camera.
First we 
prove theoretically that any candidate specular points,
for an image that is generated by a specific camera and taken under a daylight,
must lie on a straight line in log-chromaticity space, for a chromaticity that
is generated using a geometric-mean denominator.
Use is made of the assumptions that daylight illuminants can be approximated
using Planckians and that camera sensors are narrowband or can be made so by
spectral sharpening.
Then we show how a particular camera can be calibrated so as to discover this
locus. As applications 
we use this curve for illuminant detection, and also for re-lighting of images
to show they would appear under lighting having a different colour temperature.

\end{abstract}

\section{Introduction \label{SEC:CH6INTRODUCTION}}
The objective of this paper is to show that natural lights must necessarily
follow a straight-line locus, in a special 2-D chromaticity feature space
generated using a geometric-mean denominator to remove the effect of magnitude
from colour, and that this locus can be derived from a camera calibration.
Transformed back into non-log coordinates, the straight line in log colour space
means that in terms of ordinary L$_1$-norm
based chromaticity
$\{R,G\}/(R+G+B)$ lights follow a particular curve.
The locus determined is camera-dependent.
Derivation of the parameters of this locus via a camera calibration means that
then one can use the path to help identify the illuminant in the scene, and also
to transform from one illuminant to another.

In this paper, the main use we make of the above observation regarding the path following by illuminants
is to apply this additional
constraint to colour constancy algorithms as extra information that can be
brought to bear. We show that the specular-locus thus found does help in
discovering the lighting in a scene.
Moreover, since we know the path that illuminants would take depending on the colour
temperature $T$, we can re-light a scene simply by changing $T$ and thus moving
along the locus. Using measured data for changing lights for static scenes we
show below that this shift in lighting is indeed accurate.

The history of using specularities to discover the illuminant is lengthy, and
here we simply highlight some key contributions used in this paper.
Shafer~\cite{SHAFER.DICHROMATIC.85} introduced the widely used and quite effective dichromatic
model of reflectance
for dielectric materials, wherein surface reflectance consists of 
(i) a diffuse
(`body')
component that depends on subsurface material properties of a reflecting surface and 
(ii) a specular (`surface')
component that depends on the air-surface interface layer and not the body-reflectance properties.
The diffuse component is responsible for generating the colour and shading for an object and
the specular component is responsible for highlights. 
For a dielectric (e.g., plastics) the neutral-interface model
\cite{BRENEMAN.PAMI.90} states that the colour of the specular contribution is
approximately the
same as the colour of the illuminant itself.
However, simply taking specular colour as identical with light colour is
insufficient: typically, specular reflection looks white to the viewer (for
dielectric materials), but in fact a careful inspection of specular pixels shows
that the body colour is still present to some degree.

Klinker et~al.~\cite{KLINKER.IJCV.88} showed that when the diffuse colour is
constant over a surface, the colour histogram of its image forms a T-shaped distribution, with
the diffuse and specular pixels forming linear clusters. They used this information to estimate
a single diffuse colour. Therefore in order to use this principle, their
approach needed
to segment an image into several regions of homogeneous diffuse colour. 
Morever, Lee \cite{LEE.HC.ILLUM.CHROM.86} proposed a method which uses
specularity to compute illumination by
using the fact that in the CIE chromaticity diagram
\cite{WYSZECKI82}
the coordinates of the colours
from different points from the same surface will fall on a straight line
connected to the specular point.
This is the case when the light reflected from a
uniform surface is a
additive mixture of the specular component and the diffuse component.
This seminal work initiated a substantial body of work on identifying specular
pixels and using these to attempt to discover the illuminant 
\cite{LEHMANN.JOSA.01,TAN.PAMI.05}.
Another approach extending these algorithms is to define a constraint on the
possible colours of illumination, making
estimation more robust \cite{FINLAYSON.CVPR.01,FINLAYSON.IJCV.01}.

Finlayson and Drew~\cite{FINLAYSON.DREW.ICCV01} used 4-dimensional images (more
colours than R,G,B) formed
by a special 4-sensor camera. They first formed 
colour ratios to 
reduce the dimensionality to 3 and to eliminate light intensity and shading;
then projecting log values into the
plane orthogonal to the direction in the 3-D space corresponding to a lighting change
direction they arrived at generalized colour 2-vectors independent of lighting. 
They noted that
in the 2-space, specularities
are approximately linear streaks pointing to a single specular
point. Therefore they could remove specularities by the simple expedient of
replacing each 2-D colour by the maximum 2-vector position
at its particular direction from the specular point. Note, however,
that in \cite{FINLAYSON.DREW.ICCV01} 
the authors were constrained to using a {\em four-}sensor camera.
Here we relax that necessity by adding a more complete camera calibration phase. 

In \cite{CLU.FLASH.ILLEST.CIC14.2006}, Lu and Drew carried out an analysis 
again based on the formulation in \cite{FINLAYSON.DREW.ICCV01}, but in 3-D
rather than 4-D and using an
additional image generated by imaging a with-flash exposure in addition to an image with no
flash. The addition of an extra image means that by subtracting the images an
estimate of illuminant colour temperature can be established based on closeness
to a predetermined set of clusters for different lights in a log-chromaticity space,
using the mean over the image in that space compared to the clusters.

In this paper we present a new camera calibration method aimed at finding
a specular-point locus in the log-chromaticity colour feature space, for
daylight illuminants.
We prove that, in a simplifying model for image formation under
non-fluorescent illumination, any candidate illuminants
for an image generated by a specific camera 
must lie on a line in log-log chromaticity space if we use a
geometric mean to normalize colour.
This has the consequence that ordinary
$r,g$ chromaticities formed by dividing by the sum $R+G+B$ must lie on a specific curve.
To support these theoretical considerations, we demonstrate the applicability of
the 
line in log chromaticity space for several different datasets and,
as applications, we use the resulting curve for illumination recovery and
re-lighting with a different illumination.

In essence, we are proposing a type of new colour constancy algorithm, one that
uses a camera calibration.
Many colour constancy algorithms have been proposed
(see \cite{HORDLEY.CRA.06,GIJSENIJ.TIP.11} for an overview).
The foundational colour constancy method, the so-called White-Patch or Max-RGB method, estimates
the light source colour from the maximum response of
the different colour channels \cite{Land.retinex.77}. Another well-known colour
constancy method is based on the Grey-World hypothesis
\cite{Buchsbaum80}, which assumes that the average reflectance in the scene
is achromatic. Grey-Edge is a recent version of the Grey-World hypothesis
that says: the average of the reflectance differences in a scene is achromatic
\cite{vandeWeijer05}. Finlayson and Trezzi \cite{Finlayson.CIC.04} formalize
grey-based 
methods by subsuming them into a single formula using the Minkowski p-norm.
The Gamut Mapping algorithm, a more complex and more accurate algorithm, was introduced by
Forsyth \cite{FORSYTH.NOVEL.ICCV.88}. It is based on the assumption that in real-world
images, for a given illuminant one observes only a limited
number of colours. Several extensions have been proposed \cite{BARNARD.ECCV.00,FINLAYSON.HORDLEY.TIP.00,FINLAYSON.COLOR.PERSPECTIVE.96,GIJSENIJ.IJCV.08, VAEZI.ICIP.12}.

The paper is organized as follows:
To begin, in $\S$\ref{SEC:IMAGEFORMATION}
we discuss the underlying assumptions that allow us to create a
simplified model of colour image formation. Then in $\S$\ref{SEC:LOGCH} we examine how the simplified model
plus an offline calibration of the camera can be used to analyze the specular highlights. We propose 
a specular-point locus in chromaticity space in $\S$\ref{SEC:LOCUS} based on the calibration for each camera.
In $\S$\ref{SEC:ILL} and $\S$\ref{SEC:RELIGHTING} we use the proposed illuminant locus to demonstrate its
applicability in two application areas:  illuminant identification, and image re-lighting.
In $\S$\ref{SEC:MATTE} we introduce a method to generate a matte image using
our estimated illuminant, giving a specular-free image.
Finally, we conclude the paper in $\S$\ref{SEC:CONCLUSIONCH6}.

\section{Image Formation \label{SEC:IMAGEFORMATION} }
To generate a simplified image
formation model
we apply the following set of simplifying assumptions
(cf.~\cite{FINLAYSON.HORDLEY.SHADOWS.PAMI06}):
(1) illumination is Planckian or is sufficiently near the Planckian locus that a
blackbody radiator forms a reasonable approximation for this use \cite{FINLAYSON.HORDLEY.INVT.JOSA01}; (2) surfaces are dichromatic
\cite{SHAFER.DICHROMATIC.85}; and (3) RGB camera sensors are narrowband or can be
made sufficiently narrowband by a spectral-sharpening colour-space transform
\cite{FINLAYSON.SHARP.94}.

Thus we begin
by considering a narrowband camera, with three sensors.
Note again that in \cite{FINLAYSON.DREW.ICCV01} 
the authors were constrained to using a {\em four-}sensor camera.
Here we relax that necessity by adding a more complete camera calibration phase
for a camera with only three sensors.

Real camera sensor curves are not in fact narrowband:
Below, we investigate how the assumption of Planckian lighting impacts models
of image
formation by making use of a 3-sensor delta-function sensitivity camera.
It is evident that real sensors are far from idealized delta functions:
each is typically sensitive to a wavelength interval over 100nm in extent.
Nevertheless, as we shall see, they behave sufficiently like
narrowband sensors for our theory to work and moreover this behaviour could be
promoted by
carrying out calculations in an intermediate spectrally sharpened colour space
\cite{FINLAYSON.SHARP.94}.

Now let us briefly examine image
formation in general for a dichromatic reflectance function comprising Lambertian and
specular parts.
For the Lambertian component, suppose there are $i=1..L$ lights, each with 
the same SPD $E^i(\lambda)$ (e.g., an area source)
given by Wien's approximation of a Planckian source \cite{WYSZECKI82}:

\begin{equation}
E^i(\lambda) \;=\; I^i c_1 \lambda^{-5} e^{\mbox{\normalsize
		$-c_2/(\lambda T_i)$}} 
\; , \; c_1 = 3.74183\times 10^{16} \, , \, c_2 = 1.4388\times 10^{-2}
\end{equation}
with distant lighting from lights 
in normalized directions $\mba^i$ with intensities $I^i$ (the constant $c_1$
determines the units).
If the surface projecting to retinal point $\mbx$ has spectral surface reflectance
$S(\lambda)$ and normal \mbn then,
for a delta-function narrowband sensor camera with spike 
sensor sensitivities $Q_k(\lambda)$=$q_k \delta(\lambda - \lambda_k)$, $k=1..3$,
the 3-vector RGB response $R_k$ is
\begin{equation}
\begin{array}{l}
R_k \;=\; \sum_{i=1}^L \mba^i \cdot \mbn \mbox{$\int$} E^i(\lambda) S(\lambda) Q_k(\lambda) d \lambda
\\
\\
\;=\; \sum_{i=1}^L c_1 \mba^i \cdot \mbn S(\lambda_k)
I^i (\lambda_k)^{-5} e^{\mbox{\normalsize $-c_2/(\lambda_k T_i)$}}
\; q_k
\\
\\
= \; \left [ \sum_{i=1}^L (c_1 I^i \mba^i) \right ]
\cdot \mbn S(\lambda_k)
(\lambda_k)^{-5} e^{\mbox{\normalsize $-c_2/(\lambda_k
		T)$}}
\; q_k

\qquad \hfill \mbox{if all $T_i=T$}
\\
\\
\equiv \; \widetilde{\mba} \cdot \mbn \; S(\lambda_k) \lambda^{-5}
e^{\mbox{\normalsize $-c_2/(\lambda_k T)$}}
\; q_k
\vspace*{-0.15in}  
\; , 
\; k=1..3
\end{array}
\end{equation} \\

The above is the matte model employed. For the specular part,
let us assume a specular model dependent on 
the half--way vector  $\mbnS$ between 
the illuminant direction and the viewer:
\begin{equation}
\mbR^{\mbox{Specular}}
\; = \; \sum_{i=1}^L \mbb^i\!\!_S
\Phi(\mbn^i\!\!_S \, \cdot \, \mbn)
\; ,
\end{equation}
where $\mbb^i\!\!_S$ is the colour of the specularity for the $i^{\mbox{th}}$
light.
E.g., in the Phong specular model
\cite{FOLEY.VANDAM.90},
\begin{equation}
\Phi(\mbnS^i \, \cdot \, \mbn) = (\mbnS^i \, \cdot \, \mbn)^p
\quad ,
\end{equation}
where a high power $p$ makes a more focussed highlight.

Now in a neutral interface model
\cite{LEE.HC.ILLUM.CHROM.86},
the colour of the specular term is approximated as:
\begin{equation}
\mbb^i\!\!_S \; \equiv \; \mbox{colour of the light.}
\label{EQ:NIR}
\end{equation}
Hence for Lambertian plus Specular reflectance, we arrive at a simple model:
\begin{equation}
\begin{array}{ll}
R_k \; = \; 
& \left [
\widetilde{\mba} \cdot \mbn S(\lambda_k) 
\;+\; \sum_{i=1}^L c_1 I^i \Phi(\mbn^i\!\!_S \, ^T \, \mbn) 
\right ]
\lambda_k^{-5} e^{\mbox{\normalsize $-c_2/(\lambda_k T)$}}
\; q_k
\quad .
\vspace*{-0.15in}  
\end{array}
\end{equation}
For each pixel at a retinal position \mbx,
the second term in the brackets is a constant, $\beta$ say, that depends only on
geometry and not on the light colour.  Therefore we have
\begin{equation}
R_k \; = \; 
\left [
\widetilde{\mba} \cdot \mbn S(\lambda_k) 
\;+\; \beta
\right ]
\lambda_k^{-5} e^{\mbox{\normalsize $-c_2/(\lambda_k T)$}}
\; q_k
\label{EQ:ADDSPECTERM}
\end{equation}
with possibly several specular highlights on any surface ($\beta=\beta(\mbx)$).

If we define 
\begin{equation}
\alpha \; = \; \beta / ( \widetilde{\mba} \cdot \mbn  )
\quad ,
\end{equation}
then our expression simplifies to:
\begin{equation}
R_k \; =\; 
(\widetilde{\mba} \cdot \mbn)
\left [ S(\lambda_k) + \alpha \right ] \lambda_k^{-5}
e^{\mbox{\normalsize $-c_2/(\lambda_k T)$}} 
\, q_k
\label{eq:Rk}
\end{equation}

\section{Specular-Point Line in Log Chromaticity Space \label{SEC:LOGCH} }
We note that dividing by a colour channel (green, say) removes the initial factor
in eq.~(\ref{eq:Rk}).
We can divide instead by the geometric mean (cf.\ \cite{FINLAYSON.DREW.ICCV01})
so as not to be forced to choose a particular normalizing channel.
Define the mean $R_M$ by
\begin{equation}
R_M \;=\; \sqrt[3]{ \Pi_{k=1}^{3} \; R_k }
\quad .
\label{EQ:GEOMEAN}
\end{equation}
Then we can remove light intensity and shading by forming a chromaticity
3-vector \mbr via
\begin{equation}
r_k = R_k / R_M, \; k=1..3.
\label{EQ:Rk}
\end{equation}
Thus from eq.~(\ref{eq:Rk}) we have
\begin{equation}
\log r_k \;=\; \log \left (\frac{s_k + \alpha}{s_M + \alpha}\right )
\;+\; w_k
\;+\; \left ( e_k - e_M \right ) \frac{1}{T} \;, \quad  k=1..3,
\label{EQ:LOGRHOK}
\end{equation}
where we simplify the expressions by defining some short-hand notations as follows:
\begin{equation}
\begin{array}{l}
s_k \,=\,  S(\lambda_k); \; v_k \;=\; \lambda_k^{-5} q_k; \;
v_M = \left \{ \prod_{j=1}^3 \lambda_j^{-5} \, q_j \right \} ^{1/3}\; ,
w_k \;=\; \log \left ({v_k}/{v_M}\right ) \;\
\\
\\
e_k \;=\; -c_2/\lambda_k; \; 
e_M \;=\; (-c_2/3) \, \sum_{j=1}^3 (1/\lambda_j) \; , \;
\end{array}
\end{equation}
and we define an effective geometric-mean-respecting value $s_M$ by
setting
$$
(s_M + \alpha ) \;\equiv\; \left \{ \prod_{j=1}^{3} (s_j + \alpha ) \right \} ^ {1/3}
$$

In the case of broad-band sensors we replace some of the definitions in
eq.~(\ref{EQ:LOGRHOK}) above by values that are equivalent for delta-function cameras but
are appropriate for real sensors (extending definitions in
\cite{FINLAYSON.DREW.ICCV01}):
\begin{equation}
\begin{array}{l}
\sigma_k \;=\; \int q_k(\lambda) d \lambda \; , 
\\
e_k \;=\; 
(1 /  \sigma_k) \;
\int -(c_2/\lambda) q_k(\lambda) d \lambda
\; , 
\\
e_M \;=\; (1/3) \sum_{j=1}^3 e_k
\; , 
\\
s_k \;=\; 
(1 /  \sigma_k) \;
\int S(\lambda) q_k(\lambda) d \lambda
\; ,
\\
v_k \;=\; \int \lambda^{-5} \, q_k(\lambda) d \lambda
\end{array}
\end{equation}

\vspace{.15in}
The meaning of eq.~(\ref{EQ:LOGRHOK}) is that the log of the chromaticity is
given by:
(i) A term consisting of the matte-surface term $s_k$ combined with a term $\alpha$,
a scalar at each pixel that is the specular contribution;
(ii) a constant 3-vector offset term, $w_k$, which is a characteristic of the particular
camera;
and (iii) a term equal to the product of a ``lighting-change"
3-vector 
$( e_k - e_M )$, also
characterizing the camera, times the inverse of the correlated colour temperature $T$
encapsulating the colour of the light. 

Thus as the light colour (i.e., $T$) changes, say into a shadow or because of
inter-reflection, the
log-chromaticity at a pixel $\mbx$ simply follows a straight line in 3-space
(as temperature $T$ changes), along the light-change direction $( e_k - e_M )$,
even including the specular term $\alpha$.
For a fixed $T$, if $\alpha$ changes on a patch with reflectance vector $s_k$, then the plot of $\log \mbr$ will be a curved line.

In this paper, we mean to calibrate the camera so as to recover (a projection
of) both this light-change vector as well as the constant additive term $w_k$.
The difference from previous work
\cite{FINLAYSON.DREW.ICCV01}
is as follows.

In the method \cite{FINLAYSON.DREW.ICCV01}, going over to a chromaticity space meant that 4 dimensions were reduced
to 3. Then in that 3-space, light-change vector $(e_k-e_M)$ was obtained as the
first eigenvector of mean-subtracted colour-patch values. 
To then go over to a 2-space, log-chromaticity values were then projected onto the subspace orthogonal to 
3-D light-change vector. 
This meant that all lighting colour and strength were projected away.
In that plane, the illuminant, and consequently
the specular point as well, were always located in precisely the same spot.
It was argued that, at a highly specular point in an input image, the pixel
values would essentially consist of the specular point and thus one could derive
that point from training images.
Then forming radii from that specular spot out to the least-specular pixel position
effectively removed specularities.

Here, in contrast, we start with 3-D colour values, rather than 4-D ones, and so chromaticity vectors
are effectively 2-D. Now calibration of the
camera is used to provide both a value of the offset term $w_k$ in
eq.~(\ref{EQ:LOGRHOK}) as well as of the lighting-colour-change vector $(e_k-e_M)$.

For specular pixels, there is no surface term $s_k$ above in
eq.~(\ref{EQ:LOGRHOK}), and $\log(\alpha/\alpha)=0$,  so the value of this log-geometric-mean chromaticity at a
purely specular pixel becomes the simpler form
\begin{equation}
\log r_k \;=\; 
w_k
\;+\; \left ( e_k - e_M \right ) \frac{1}{T} \;, \quad  k=1..3,
\label{EQ:SPECLOGRHOK}
\end{equation}

Thus as $T$ changes we have a line, in a 2-D colour space, whereon any specular point must lie.
To determine just where it does lie, we form an objective function measure, which is in
fact minimized provided we choose the correct value of $T$: an example of such a
measure is given below in $\S$\ref{SUBSEC:PLANE}.
Hence we recover
the temperature $T$ and therefore the light colour.
Moreover, since have an illuminant locus we can go on to re-light images by
moving the illuminant along the locus obtained during the camera calibration
phase. Such re-lit images are shown below in $\S$\ref{SEC:RELIGHTING}
where images are shown as they would appear under a
different colour temperature.

Note that although we work with 3-vectors, the step of division by the geometric
mean creates $\log \mbr$ vectors that lie on a plane:  they are all orthogonal
to the vector $(1,1,1)^T$ --- in fact, each of the three terms in
eq.~(\ref{EQ:LOGRHOK}) lies in this plane.
Thus the components are not independent. 

\section{Recovery of Specular-Point Locus \label{SEC:LOCUS}}
To find the vector $(e_k-e_M)$, $k=1..3$ we image matte Lambertian colour patches.
Here we use the 18 non-grey patches of the 
Macbeth ColourChecker \cite{MCCAMY.MACBETH.76}.
We form $\log r_k$ values using temperatures $T$ from $5500^\circ$K
to $10500^\circ$K.

According to eq.~(\ref{EQ:LOGRHOK}) (with no specular contribution), 
for each surface we should see a set of
points in 3-space that falls on a straight line along $(e_k-e_M)$.
Thus for each surface, if we then subtract the mean, in each channel $k$ of
$\log r_k$, we see a set of nearly coincident lines through the origin.

Therefore, 
as pointed out in \cite{FINLAYSON.HORDLEY.INVT.JOSA01} 
(in a 2-D setting like eq.(\ref{EQ:LOGRHOK}) but with $k=1..2$),
we can find vector $(e_k-e_M)$ by forming the covariance matrix of
mean-subtracted 
$\log r_k$
values and calculating eigenvectors.  The first eigenvector is the
desired approximation of direction $(e_k-e_M)$.

To derive the offset term $w_k$, we utilize the recovered normalized version of
vector $(e_k-e_M)$ and image two
lights (below) to determine the scaling along the inverse-temperature line.

Since we know that our colour features lie on the plane perpendicular to the
unit vector $\mbu=1/\sqrt{3} (1,1,1)^T$, to simplify the geometry
we first rotate all our 
log-chromaticity vector coordinates into that plane by forming the projector $P_u$ onto the \mbu
direction.  2-D coordinates $\mbchi$ are formed 
by multiplication of the rotation matrix $U$ from the eigenvector
decomposition of the projector $P_u^\bot=I-P_u$ onto the plane:
\begin{equation}
P_u^\bot \;=\; U^T \; diag(1,1) \; U, \qquad \mbox{$U$ is $2\times 3$}.
\end{equation}
We denote 2-vectors in this 2-D space as \mbchi. And, explicitly, we form 2-vectors in the plane by
\begin{equation}
\mbchi \;=\; U \, \log\mbr
\label{EQ:PROJECT3TO2}
\end{equation}

Now suppose that in the 2-D coordinates \mbchi, two lights $E_1$ and $E_2$ produce
vectors $\mbchi^{E_1}$ and $\mbchi^{E_2}$: for each light we form chromaticity (\ref{EQ:Rk}),
take logs, and then
project via (\ref{EQ:PROJECT3TO2}).
Consider the recovered {\em normalized} light-change direction vector,
projected into this plane: define the 3-vector \mbe as having components
$(e_k-e_M)$, and denote
its unit 2-vector projection as $\hat{\mbxi}$.  
Note that we recover only a {\em normalized} version of $\mbe$ from our SVD
analysis of imaged colour patches, with the norm unknown. That is, we work in the
plane by rotating with $U$, and further normalize that projected 2-vector, giving
a known, normalized, 2-vector $\hat{\mbxi}$ from our calibration.

Also, denote by \mbeta the projected
vector $w_k$: this is what we aim to recover.
$$
\begin{array}{l}
\hat{\mbxi} \;=\; ( U \mbe ) / \nu \, , \;\; \mbox{where} \;\; \nu \; \equiv \; \|U \mbe \|,
\\
\mbeta \;=\; U \mbw \, .
\end{array}
$$
Then the 2-vector coordinates for the two lights $E_i,\, i=1..2$ are
\begin{equation}
\chi^{E_i}_\mu \;=\; \eta_\mu \,+\, \nu \hat{\xi}_\mu/T_i , \; i=1..2,\, \mu=1..2
\end{equation}
where $\nu$ is an unknown scale, and $T_i$ are known colour-temperatures. 
Note that since we are imaging lights, not surfaces, the surface term $s_k$ in
eq.~(\ref{EQ:LOGRHOK}) is not present.

Forming the difference 2-vector $(\mbchi^{E_2}-\mbchi^{E_1})$, we obtain 
a result involving only
the normalized direction $\hat{\mbxi}$. So we can determine the norm
$\nu$ if we know $T_1$ and $T_2$.  
For consider the difference 2-vector
\begin{equation}
\chi^{E_1}_\mu \,-\, \chi^{E_2}_\mu \;=\;  \nu \hat{\xi}_\mu \; 
\left (  \frac{1}{T_1} - \frac{1}{T_2}  \right )
\label{EQ:GETk}
\end{equation}
Even from these two data points we can easily determine the normalized vector
$\hat{\mbxi}$ since it is simply given by the direction of the
difference in $\mbchi$. Since we know $T_1,T_2$, the norm $\nu$ thus falls out of
eq.~(\ref{EQ:GETk}).

Finally, subtracting the term $\nu \hat{\xi}_\mu/T_i$, $i=1,2$,
from each of the two $\mbchi$ vectors and taking the mean, we recover the offset term \mbeta. 

Let us denote by \mbxi the product $\nu \hat{\mbxi}$,
so using this vector and the
offset \mbeta we arrive at a line (for this particular camera calibrated as above)
parametrized by temperature $T$
that must necessarily be traversed by any candidate specular point:
\begin{equation}
\mbchi \;=\; \mbeta \,+\, (1/T)\mbxi 
\label{EQ:LINE}
\end{equation}

In summary, the calibration algorithm proposed is expressed in algorithm~\ref{AL:CH6_1}.

\begin{algorithm}[ht]
	\caption{ Proposed Camera Calibration }
	\label{AL:CH6_1}
	\begin{tabbing}
		S\= for \= for \= \kill
		\>  Colour target:\\
		\>\> \it Record RGB responses $R_k$, $k=1..3$ (reflected from colour target)
		\\\>\>\> \it \underline{for several lights} $\longrightarrow$ each pixel
		follows a parallel straight line;
		\\\>\>\> \it calculate geometric mean at each pixel from eq.~(\ref{EQ:GEOMEAN}).\\
		\>\> \it Derive geometric-mean-based chromaticity 3-vector \mbr from eq.~(\ref{EQ:Rk}),
		and take logarithms.\\
		\>\> \it Find 3-vector $(e_k-e_M)$ as first eigenvector for $\log \mbr$
		values, mean-subtracted 
		\\\>\>\> \it for each colour patch.\\ \\
		\>For illuminants $E(\lambda)$, characterized by their known temperatures
		$T$ 
		\\\>\>\>(in a light-box, for example):\\
		\>\> \it Derive $\log \mbr$ as above, for light reflected from a grey patch.\\
		\>\> \it Project $\log \mbr$ onto plane orthogonal to $(1,1,1)$ via
		eq.~(\ref{EQ:PROJECT3TO2}), forming 2-D coordinates $\mbchi$.\\
		\>\> \it Subtracting pairs of $\mbchi$ values for known values $T$, find 2-D
		projected light-change 
		\\\>\>\> \it vector \mbxi  via eq.~(\ref{EQ:GETk}).\\
		\>\> \it Using $\mbxi$, find mean value of camera offset vector \mbeta over \mbchi vectors 
		used, eq.~(\ref{EQ:LINE}).\\
	\end{tabbing}
	\vspace{-0.25in}
\end{algorithm}

As set out in algorithm~\ref{AL:CH6_1}, a more accurate way to recover the offset term \mbeta
and the vector \mbxi is to utilize several different known illuminants
and capture them using  the camera to be calibrated: lights should approximately 
lie on a straight line in \mbchi space. Then line parameters \mbeta and
\mbxi,
as well as outliers, can be recovered using a robust regression method such as the
Least Median of Squares (LMS)  \cite{ROUSSEEUW.BOOK.87}.

We shall find in the following sections that the offset \mbeta and the vector 
\mbxi are all the calibration information that
we need for different applications such as illuminant identification and re-lighting.

\subsection{Real Images \label{SUBSEC:REAL} }

\begin{figure}[htbp]
	\begin{center}
		\begin{tabular}[t]{c}
			\includegraphics[width=5.5cm]{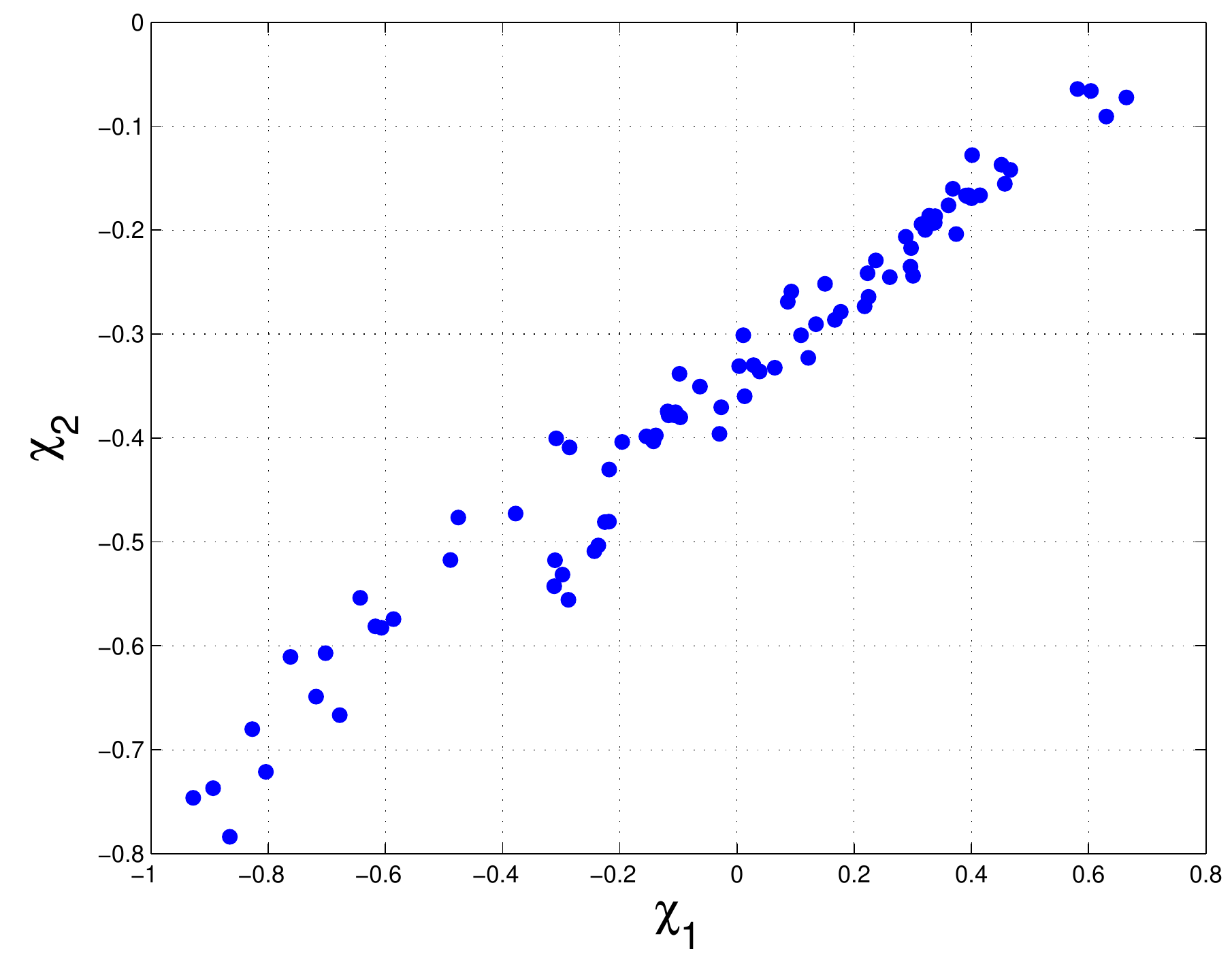} 
		\end{tabular}
		\vspace*{-0.3in}
		\caption{
			86 illuminants for Canon camera in \mbchi space \cite{GEHLER.CVPR08}. 
			Note that these illuminants do approximately follow a straight line.
		}
		\label{FIG:XLINE}
	\end{center}
	\BUT
\end{figure}

The image formation theory used is based on three idealized assumptions: (1) Planckian illumination, (2) dichromatic surfaces; and (3) narrowband camera sensors.
To determine if real images stand up under these constraints and generate the
needed straight line in 2-D colour space, we make use of datasets of measured
images  \cite{GEHLER.CVPR08,KOBUS.DATASET.02}.
Fig.~\ref{FIG:XLINE} displays measured illuminant points in \mbchi space for 86
scenes
captured by a high-quality Canon DSLR camera for 86 different lighting
conditions \cite{GEHLER.CVPR08}. Notwithstanding the fact that the camera
sensors are not narrowband and illuminants are not perfectly Planckian, we can see that these illuminants 
do indeed approximately form a straight line, thus justifying the suitability of
the theoretical formulation.

\begin{figure}[htbp]
	\begin{center}
		\begin{tabular}[t]{c}
			\subfigure[]{\includegraphics[width=5.5cm]{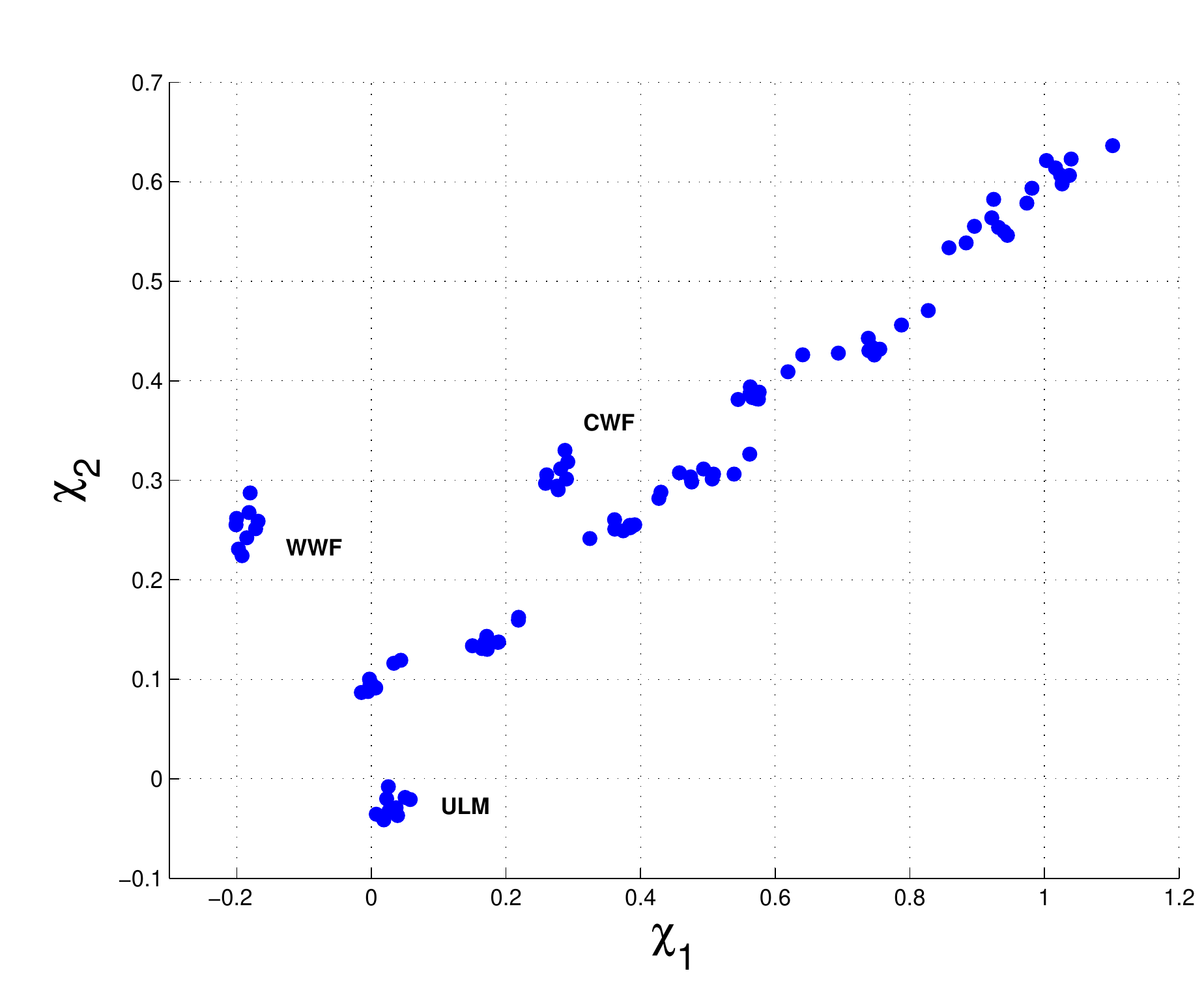} } 
			\subfigure[]{\includegraphics[width=5.5cm]{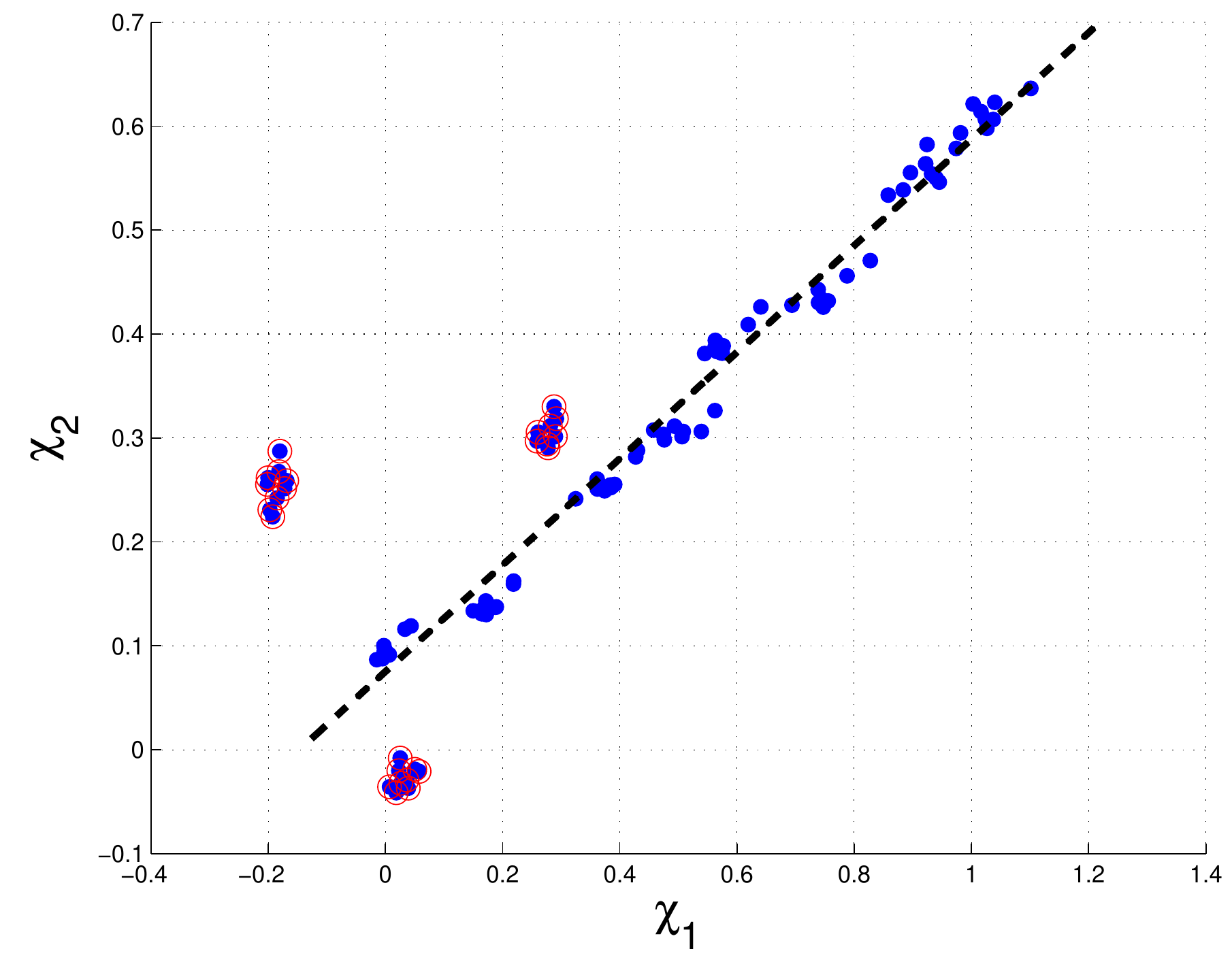} } 
		\end{tabular}
		\vspace*{-0.25in}
		\caption{
			(a):\ 98 illuminants for images containing significant specular content
			\cite{KOBUS.DATASET.02},
			plotted in 2-D \mbchi colour space. Note clusters of points that arise from
			fluorescent illuminants (WWF, CWF, ULM).
			(b):\ Outliers automatically determined by LMS regression are shown using a red circle,
			and the regression line is shown as black dashed.
		}
		\label{FIG:KOBUS98LINE}
	\end{center}
	\BUT
\end{figure}

Since we assume that lights can be characterized as Planckian, we expect that 
severely non-Planckian lights will form outliers to the straight-line path
determined.
Figs.~\ref{FIG:KOBUS98LINE}(a,b) demonstrate that this is indeed that case.
Here we show illuminant points transformed to 2-D \mbchi space for 98 images
consisting of measured images of 9 objects that are specifically selected to
include substantial specular content, under different illumination conditions \cite{KOBUS.DATASET.02}. 
In this dataset, illuminants for 26 of the images are fluorescent (Sylvania Warm White Fluorescent(WWF), Sylvania Cool White Fluorescent (CWF) and Philips Ultralume Fluorescent(PUF) ).
These show up in
Fig.~\ref{FIG:KOBUS98LINE}(a) as outlier points. Fig.~\ref{FIG:KOBUS98LINE}(b)
shows that the robust LMS method correctly identifies these points as outliers
and thus does not include them in calculating line parameters.

\section{Illuminant Identification \label{SEC:ILL}}

Our camera calibration process has generated a locus in chromaticity space that
candidate natural daylight illuminations will follow. 
In this section we show how for a new image we can identify a point on this locus as an
estimate of the illuminant.

Recently, Drew et~al.\ \cite{DREW.CPCV.2012, DREW.CVIU.2014} presented an illuminant
estimation method based on a planar constraint. This stated that
for near-specular pixels, Log-Relative-Chromaticity (LRC) values are orthogonal
to the light chromaticity:
they showed that if one divides image chromaticity by illuminant chromaticity,
then in a log space the resulting set of 3-vectors are approximately planar, for
near-specular pixels, and orthogonal to the lighting --- for the correct choice
of the illuminant only.
Hence they propose an objective function based on this
planar constraint which is minimized for the correct illuminant.

Here, we utilize this daylight illuminant planar constraint by further
constraining the light to lie on the
daylight locus we have derived above.  The locus provides an additional
constraint on the illuminant and hence improves the estimate.

To begin, we briefly recapitulate below the derivation of this planar constraint.

\subsection{Plane Constraint \label{SUBSEC:PLANE}}

Suppose we rewrite eq.~(\ref{eq:Rk}) for the 3-vector RGB response $\mbR$, here
relinquishing the requirements that lighting be Planckian and sensors be
narrowband, but instead applying the different simplifying assumption 
that matte pixel 3-vector RGB triples be a component-wise product of a light
3-vector $epsilon_k$, $k=1..3$, and a surface triple $\varsigma_k$ \cite{BORGES.JOSA.91}.
Here, $\varsigma_k$ is the  reflectance at a pixel under equi-energy white light.

Adding a Neutral Interface Model term \cite{LEE.HC.ILLUM.CHROM.86} for specular content, as in
eq.~(\ref{EQ:ADDSPECTERM}), we have approximately
\begin{equation}
R_k \; \simeq \; \frac{\kappa \varsigma_k \epsilon_k}{q_k} \;+\; \beta \epsilon_k
\end{equation}
where $\kappa$ is shading. E.g., for Lambertian matte shading
$\kappa$ equals
lighting-direction dotted into surface normal.
Here, $q_k$ is again a triple giving the overall camera sensor strength
\cite{DREW.SPECTRAWITHOUTSPECTRA.JOSA03};
$\beta$ represents the amount of specular
component at that pixel. The value of $\beta$ for a pixel will depend upon the
lighting direction, the surface normal, and the viewing geometry \cite{SHAFER.DICHROMATIC.85}.
Let us lump values $\kappa \varsigma_k/q_k$ into a single quantity and for convenience
call this simply $\varsigma_k$.
Now we have 
\begin{equation}
R_k \; = \; \varsigma_k \epsilon_k \;+\; \beta \epsilon_k
\end{equation}

Instead of the geometric-mean based chromaticity $\mbr$ in eq.~(\ref{EQ:Rk}),
let us make use of the standard L$_1$-norm based chromaticity \cite{WYSZECKI82} 
\begin{equation}
\mbrho \;=\; \{R,G,B\}/(R+G+B)
\label{EQ:RHOk}
\end{equation}
Thus here we have
\begin{equation}
\rho_k \;=\; \frac{\varsigma_k \epsilon_k \;+\; \beta \epsilon_k}{\sum_{j=1}^3
	( \varsigma_j \epsilon_j \;+\; \beta \epsilon_j ) }
\vspace{.1in}
\end{equation}

Let us define the Log-Relative-Chromaticity (LRC) as the above chromaticity
divided by the chromaticity for the lighting itself, $\rho^\epsilon_k$.
The planar constraint \cite{DREW.CPCV.2012} says that for near-specular pixels,
LRC values are orthogonal to the light chromaticity, provided we have chosen the
correct illuminant to divide by.

To see how this constraint arises, form the LRC, which we denote as $\psi_k$:
\begin{equation}
\psi_k \;=\; \log \left( \frac{\rho_k}{\rho^\epsilon_k} \right) \;=\;  \log
\left( \frac{\varsigma_k \epsilon_k \;+\; \beta
	\epsilon_k}{\sum_{j=1}^3 ( \varsigma_j \epsilon_j ) \;+\; \beta \sum_{j=1}^3 \epsilon_j } \cdot
\frac{\sum_{j=1}^3 \epsilon_j}{\epsilon_k} \right) \;=\; \log \left(
\frac{\varsigma_k + \beta}{ \frac{(\sum_j \varsigma_j \epsilon_j)}{E} + \beta } \right)
\end{equation}
For convenience, now define $E \equiv \sum_{j=1}^3 \epsilon_j = |\mbepsilon\!|$ where $|\cdot|$ is the L$_1$
norm. 

Near a specular point, we can take the limit as $(1/\beta) \rightarrow 0$.
Let $\alpha = 1/\beta$. Then in the limit, $\mbpsi$ goes to
\begin{equation}
\begin{array}{lll}
\psi_k & = & \lim_{\alpha \rightarrow 0}\; \log \left \{ 
\left ( \alpha \varsigma_k + 1 \right )
/
\left ( \alpha \sum_j(\varsigma_j \epsilon_j)/E +1 \right )
\right \} \; \simeq \; \alpha \left ( \varsigma_k -  \frac{\sum_j \varsigma_j \epsilon_j} { E } \right )
\end{array}
\vspace{.1in}
\end{equation}

The above is the Maclaurin series, accurate up to $O\left ( \alpha^2 \right )$.
By inspection, we have that the LRC vector,
$\psi_k$, is {\em orthogonal to the illuminant vector:} $\sum_{k=1}^3 \psi_k e_k = 0$,
and hence also orthogonal to the illuminant chromaticity, 
$\sum_{k=1}^3 \psi_k \rho^e_k = 0$.

The planar constraint therefore suggests finding which illuminant amongst
several candidates is the correct choice, for a particular image, by minimizing
the dot-product over illuminants, for pixels that are likely near-specular
\cite{DREW.CPCV.2012}. Define $\zeta$ as the dot-product between $\mbzeta$
and the chromaticity for a candidate
illuminant, with \mbzeta formed by dividing by this same
illuminant chromaticity:
\begin{equation}
\zeta \;=\; - \mbpsi \cdot \mbrho^e \;=\; -\log(\mbrho / \mbrho^e) \cdot \mbrho^e
\label{EQ:CH6ZETAIMAGE}
\end{equation}
Then we seek to solve an optimization as express in algorithm~\ref{AL:CH6_2}
\begin{algorithm}[ht]
	\caption{ Illumination Estimation by Zeta using Optimization }
	\label{AL:CH6_2}
	\begin{eqnarray}
	Minimize & \min_{\mbrho^e} \; \sum_{\mbpsi \in \Psi_0} |\zeta|   \nonumber \\
	\mbox{subject to} & \sum_{k=1}^3{\rho_k^e}=1 \, ,  
	\, 0<\rho_k^e<1 \, , \, k=1..3 
	\label{EQ:CH6OPT}
	\end{eqnarray}
	
\end{algorithm}

where $\Psi_0$ is a set of pixel dot-product values
with the candidate illuminant chromaticity $\mbrho^e$ that are likely to be near
specular, e.g.\ those in the lowest 10-percentile.

To include the Daylight Locus constraint, for a camera calibrated as above in algorithm~\ref{AL:CH6_2}, we
consider only natural illuminants lying on the curve (\ref{EQ:LINE}).

\subsection{Experimental Results}

We apply our proposed method to two different real-image datasets 
\cite{KOBUS.DATASET.02,CARDEI.CIC.2003}
and compare our results to other colour constancy algorithms.
The motivation here is to investigate whether the derived daylight locus
correctly helps identify illuminants that are indeed daylights. We show that
this is the case.

\subsubsection{Laboratory Images}
Our first experiment uses the Barnard dataset~\cite{KOBUS.DATASET.02}, denoted
here as
the SFU Laboratory dataset (introduced above in $\S$\ref{SUBSEC:REAL}). This contains 321 measured images under 11 different measured illuminants.
The scenes are divided into two sets as follows:
minimal specularities (22 scenes, 223 images -- i.e., 19 missing images); and
non-negligible dielectric specularities (9 scenes, 98 images  -- 1 illuminant is
missing for 1 scene). In this dataset the illuminant for 86
of the images
are fluorescents.
To compare to other colour constancy methods, we consider the following
algorithms: White-Patch, Grey-World, and Grey-Edge implemented by \cite{vandeWeijer05}.
For Grey-Edge we use optimal settings, which differ per dataset~\cite{GIJSENIJ.WEBSITE} 
($p=7$ , $\sigma=4$ for the SFU Laboratory dataset and $p=1$, $\sigma=1$ for the
GreyBall dataset below). 
We also show the results
provided by Gijsenij et~al.\ \cite{GIJSENIJ.IJCV.08} for pixel-based gamut
mapping, using the best gamut mapping settings for each dataset.

How the daylight locus information is used is as an additional constraint to
the optimization (\ref{EQ:CH6OPT}), whereby candidate illuminants are restricted to
the daylight locus determined by our calibration, for the camera used in taking
images.

Table~\ref{TB:CH6KOBUS} lists the accuracy of the proposed method for the SFU Laboratory dataset 
\cite{KOBUS.DATASET.02}, in terms of the mean and median
of angular errors, compared to other colour constancy algorithms applied to this dataset. Since the daylight locus is 
designed for natural lights (Planckian illuminants) and
not fluorescents, we 
expect performance to be better for non-fluorescents, and this is indeed the
case for the 86 scenes imaged under fluorescent lighting.
As well, we break out results for all methods for non-fluorescent illuminants
(235 images). The results show that in fact using the daylight locus outperforms
all other methods in terms of median error, notwithstanding the fact that it is
a much less complex method than the gamut-mapping algorithms and does not require any
tuning parameters. 

The main conclusion to be drawn from this experiment is that the
daylight locus does aid a planar-constraint driven illuminant identifier when
illuminants are indeed natural lights.
This justifies the suitability of our daylight-locus formulation as a useful
physics-based constraint of natural lighting.

 \BUT
\begin{table} 
	\caption{Angular errors for several colour constancy algorithms, for SFU Laboratory dataset
		\cite{KOBUS.DATASET.02}.}
	\label{TB:CH6KOBUS}
	\vspace{-0.25in}
	\begin{center}
		\begin{tabular}{|l|c|c|c|c|}
			\hline
			& \multicolumn{2}{c|}{all} & \multicolumn{2}{c|}{non-fluorescent} \\
			
			Method & Median Er & Mean Er & Median Er & Mean Er  \\
			\hline\hline
			White-Patch & 																											$6.5^{\circ}$		& $9.1^{\circ}$        & $6.9^{\circ}$		& $9.9^{\circ}$\\
			Grey-World  & 																											$7.0^{\circ}$  	& $9.8^{\circ}$        & $6.4^{\circ}$		& $9.4^{\circ}$\\
			Grey-Edge ($p=1$,$\sigma=6$) & 																			$3.2^{\circ}$ 	& $5.6^{\circ}$        & $2.9^{\circ}$		& $5.3^{\circ}$\\
			Gamut Mapping pixel ($\sigma=4$)  & 												      	$2.3^{\circ}$ 	& $3.7^{\circ}$        & $1.8^{\circ}$		& $3.5^{\circ}$\\
			Planar Constraint Search & 																	        $1.9^{\circ}$   & $4.3^{\circ}$        & $1.9^{\circ}$		& $4.6^{\circ}$\\
			{\bf Daylight Locus using Planar Constraint }  & 										$2.4^{\circ}$   & $5.1^{\circ}$        & {\bf 1.6$^{\circ}$}		& $4.4^{\circ}$\\
			\hline
		\end{tabular}
	\end{center}
    \BUT
\end{table}

\subsubsection{Real-World Images}
For a more real-world (out of the laboratory) image experiment we used GreyBall
dataset provided by Ciurea and Funt \cite{CARDEI.CIC.2003}: this dataset
contains 11346 images extracted from
video recorded under a wide variety of imaging
conditions. The images are divided into 15
different clips taken at different locations. The ground truth
was acquired by attaching a grey sphere to the camera, 
displayed in the bottom-right corner of the image. This grey sphere must be
masked during experiments.

\begin{figure}[htbp]
	\begin{center}
		\begin{tabular}[t]{c}
			\subfigure[]{\includegraphics[width=7cm]{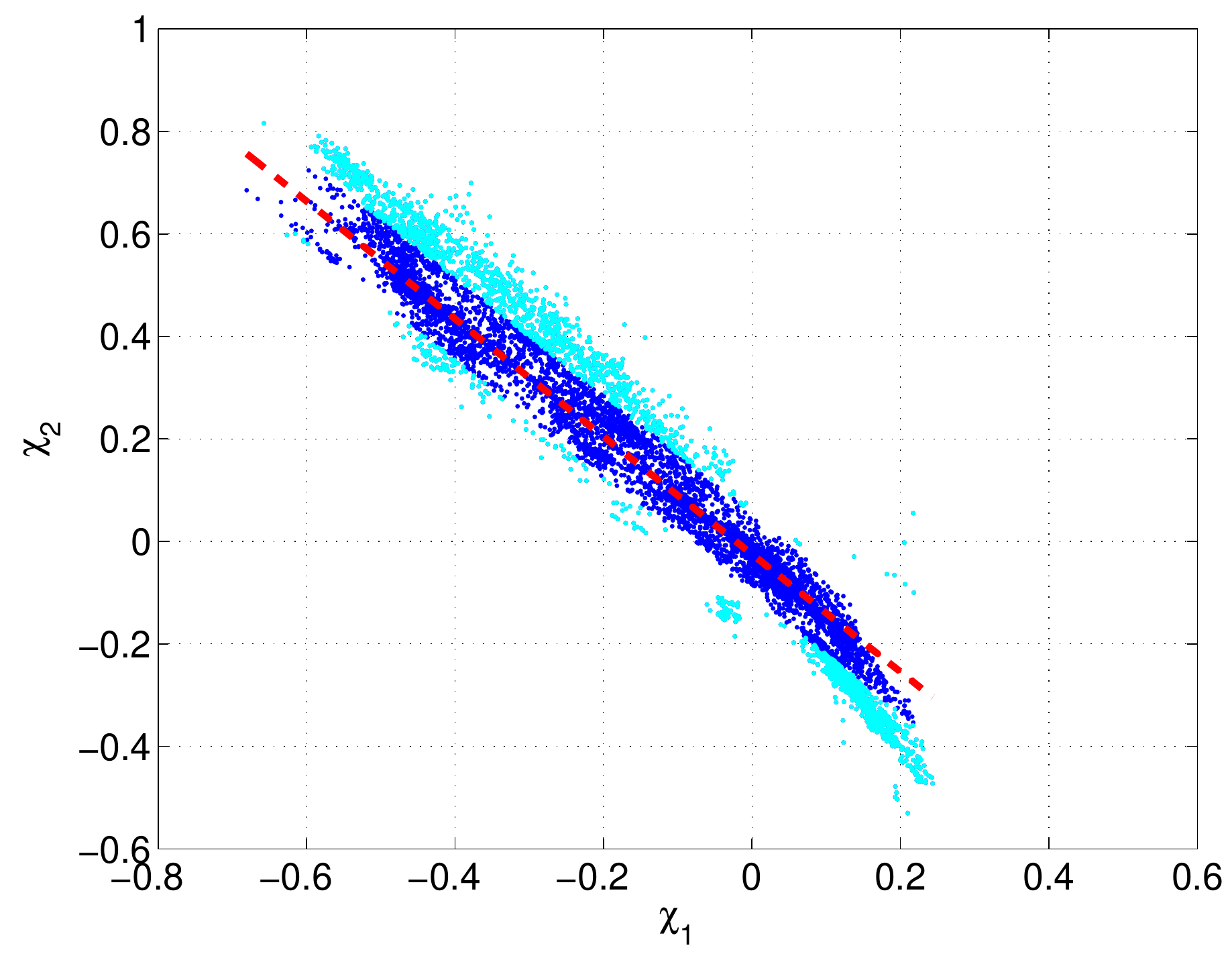} } 
			\subfigure[]{\includegraphics[width=7cm]{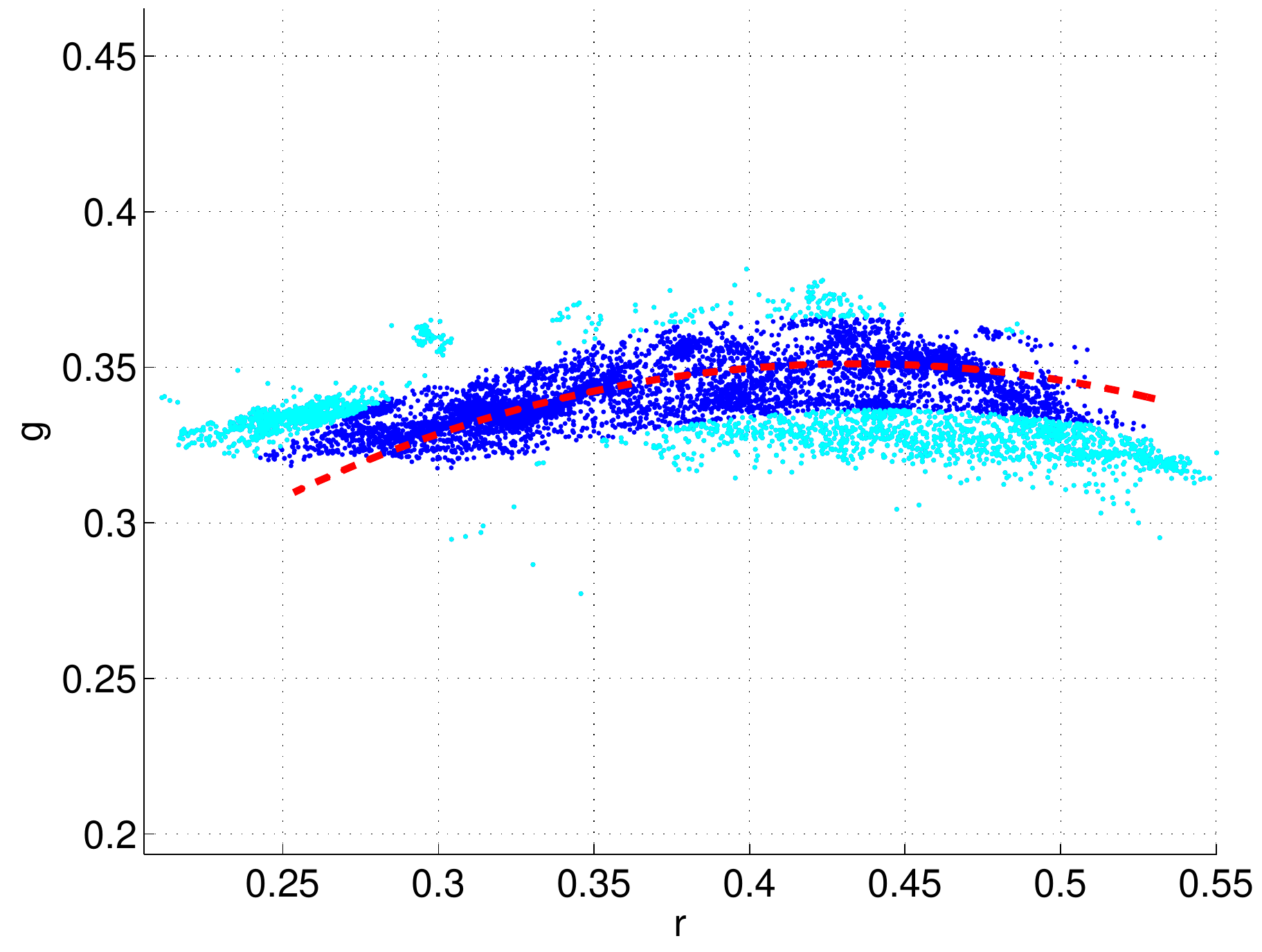} } 
		\end{tabular}
	\vspace{-0.25in}
		\caption{
			(a):\ 11346 illuminants of GreyBall data set \cite{CARDEI.CIC.2003} in \mbchi
			2-space: they approximately follow a straight line locus.
			(b):\ The illuminants transformed back into a curve in L$_1$-norm based chromaticity space. }
		\label{FIG:GREYBALLILL}
	\end{center}
	\BUT
\end{figure}

Fig.~\ref{FIG:GREYBALLILL}(a) shows the illuminants for this image set, mapped
into 2-D \mbchi colour space eq.~(\ref{EQ:PROJECT3TO2}). We see that 
these illuminants do approximately follow a straight-line path in 2-space; the
LMS-based robust regression method finds a straight-line regression line shown
red-dashed.
Transformed back into standard L$_1$-norm based chromaticity space
(\ref{EQ:RHOk}) the path is curved, as in Fig.~\ref{FIG:GREYBALLILL}(b).

Table~\ref{TB:CH6GRAYBALL} shows results for this dataset.
We find that the Daylight Locus using the Planar Constraint does better
than all the other methods save one: it is only bested by 
the far more complex Natural Image Statistics method \cite{GIJSENIJ.PAMI.11}.
This is a machine learning technique to select and combine a set of colour
constancy methods based on natural image statistics and scene semantics.
Again, we find that adding the Daylight Locus information improves the Planar
Constraint approach since here lights used are natural daylights.

\BUT
\begin{table} 
	\caption{Angular errors for several colour constancy algorithms for GreyBall dataset \cite{CARDEI.CIC.2003}.}
	\vspace{-0.25in}
	\label{TB:CH6GRAYBALL}
	\begin{center}
		\begin{tabular}{|l|c|c|}
			\hline
			Method & Median Er & Mean Er  \\
			\hline\hline
			White-Patch & 																											$5.3^{\circ}$		& $6.8^{\circ}$ \\
			Grey-World  & 																											$7.0^{\circ}$  	& $7.9^{\circ}$\\
			Grey-Edge ($p=1$,$\sigma=1$) & 																			$4.7^{\circ}$ 	& $5.9^{\circ}$ \\
			Gamut Mapping pixel ($\sigma=4$)  & 													$5.8^{\circ}$ 	& $7.1^{\circ}$ \\
			Natural Image Statistics~\cite{GIJSENIJ.PAMI.11}	& 	                    									$3.9^{\circ}$ 	& $5.2^{\circ}$ \\
			Planar Constraint Search & 																	  $4.6^{\circ}$   & $5.9^{\circ}$ \\
			{\bf Daylight Locus using Planar Constraint }  &
			{\bf 4.1}$^{\circ}$   & $5.6^{\circ}$  \\
			
			\hline
		\end{tabular}
	\end{center}
\end{table}


\section{Re-Lighting Images \label{SEC:RELIGHTING}}

We have shown that by means of calibrating the camera we can recover the
specular point for a new image not in the calibration set.
That is, 
the method recovers an estimate of the temperature $T$ for the actual
natural illuminant in a test image.
Moreover, we
have a curve that illuminants must traverse as the lighting colour changes.
Consequently 
it should be possible to re-light an image by changing the position
of the specular point along the curve, thus generating new images with a different illuminant.

If we again adopt the assumption that 
camera sensors are narrow-band, we can use a diagonal 
colour space transform \cite{CHONG.ICCV.07}
to move the image into new light conditions, via the following equation:
\begin{equation}
\begin{array}{l}
\mbM \;=\; diag(\mbrho^{e'}) \; diag(\mbrho^e)^{-1} 
\\
\\
\mbR '=\mbR \, \mbM
\end{array}
\label{eq:diagTransform}
\end{equation}
where $ diag(\mbrho^e)$ is a $3\times 3$ diagonal matrix with values from vector
$\mbrho^e$, and with $\mbrho^e$ the
current specular point and $\mbrho^{e'}$ the new specular point;  $\mbR$ and $\mbR '$
are the original and transformed $RGB$ vectors for each image pixel. 

\begin{figure*}[htbp]
	\begin{center}
		\begin{tabular}[t]{c}
			\subfigure[]{\includegraphics[width=3.75cm]{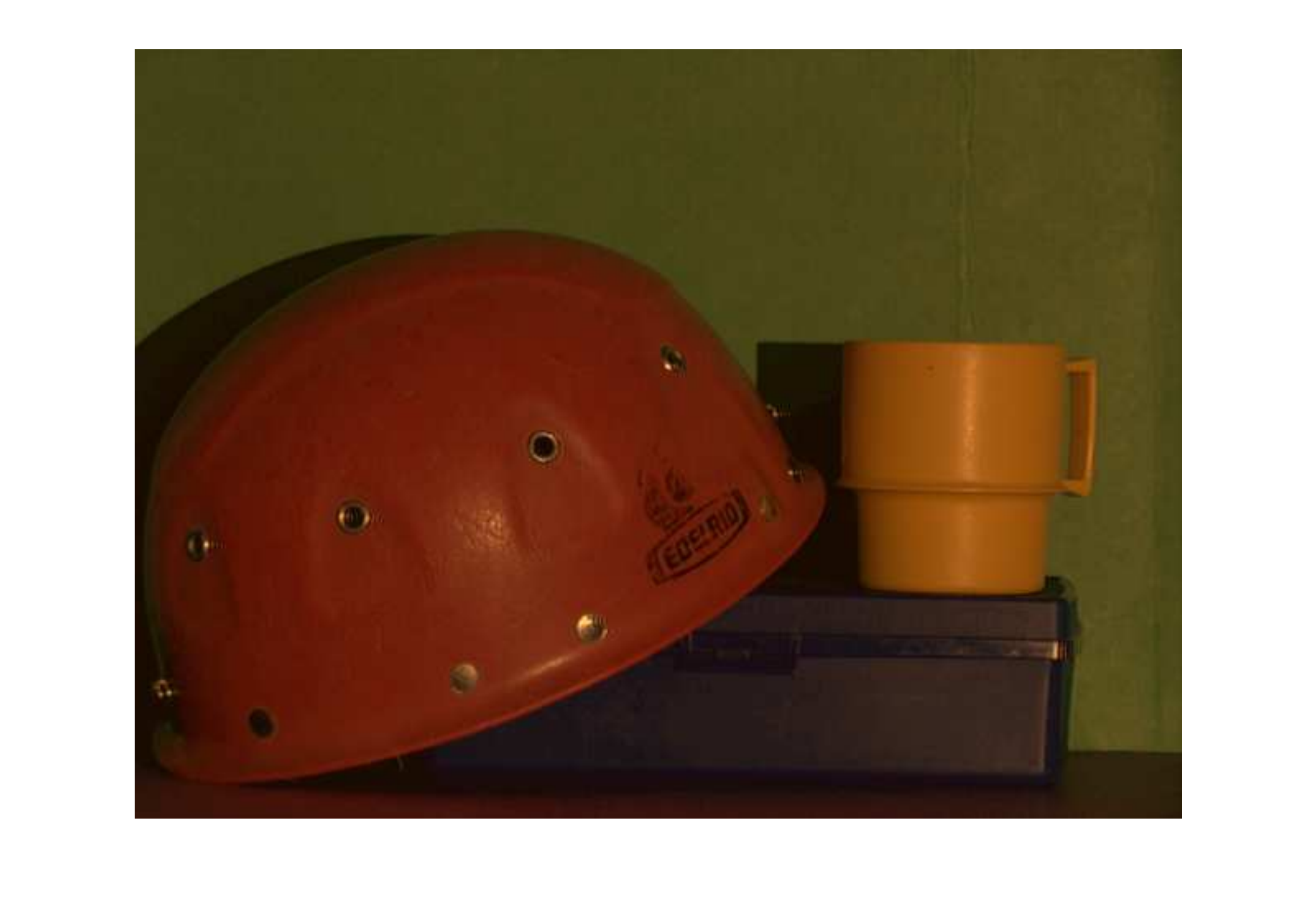}\ }
			\subfigure[]{\includegraphics[width=3.75cm]{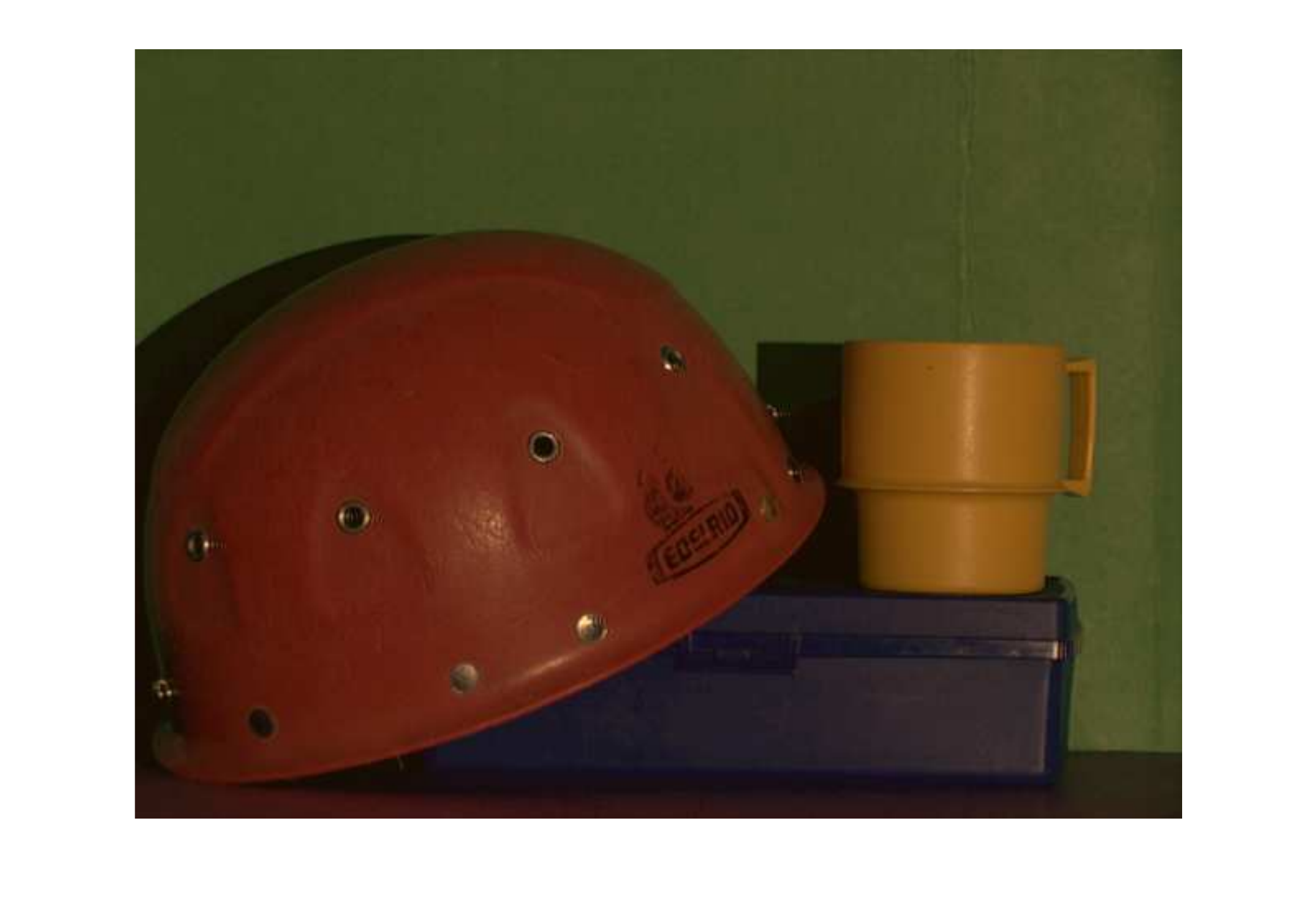}\ }
			\subfigure[]{\includegraphics[width=3.75cm]{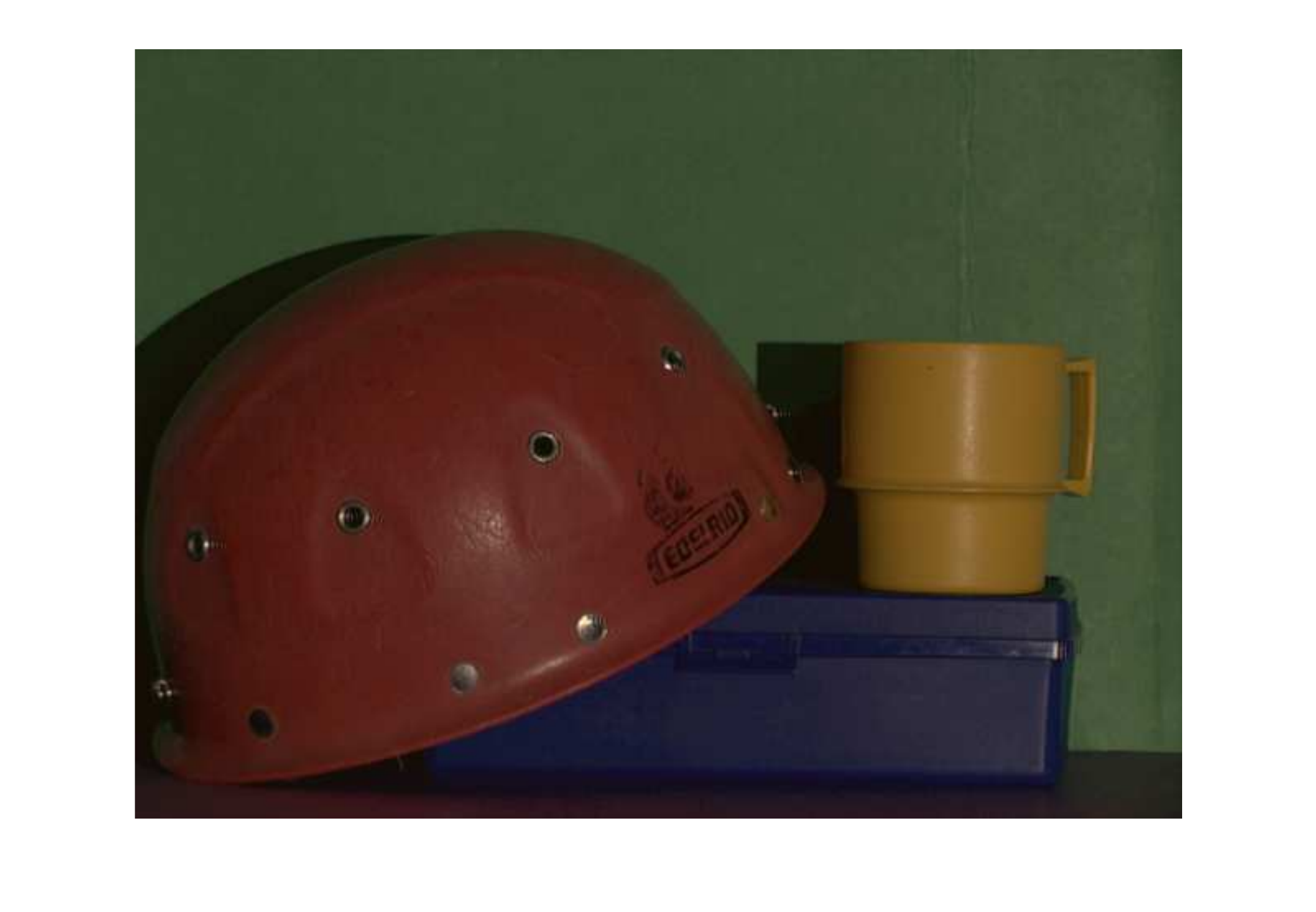}\ }
			\subfigure[]{\includegraphics[width=3.75cm]{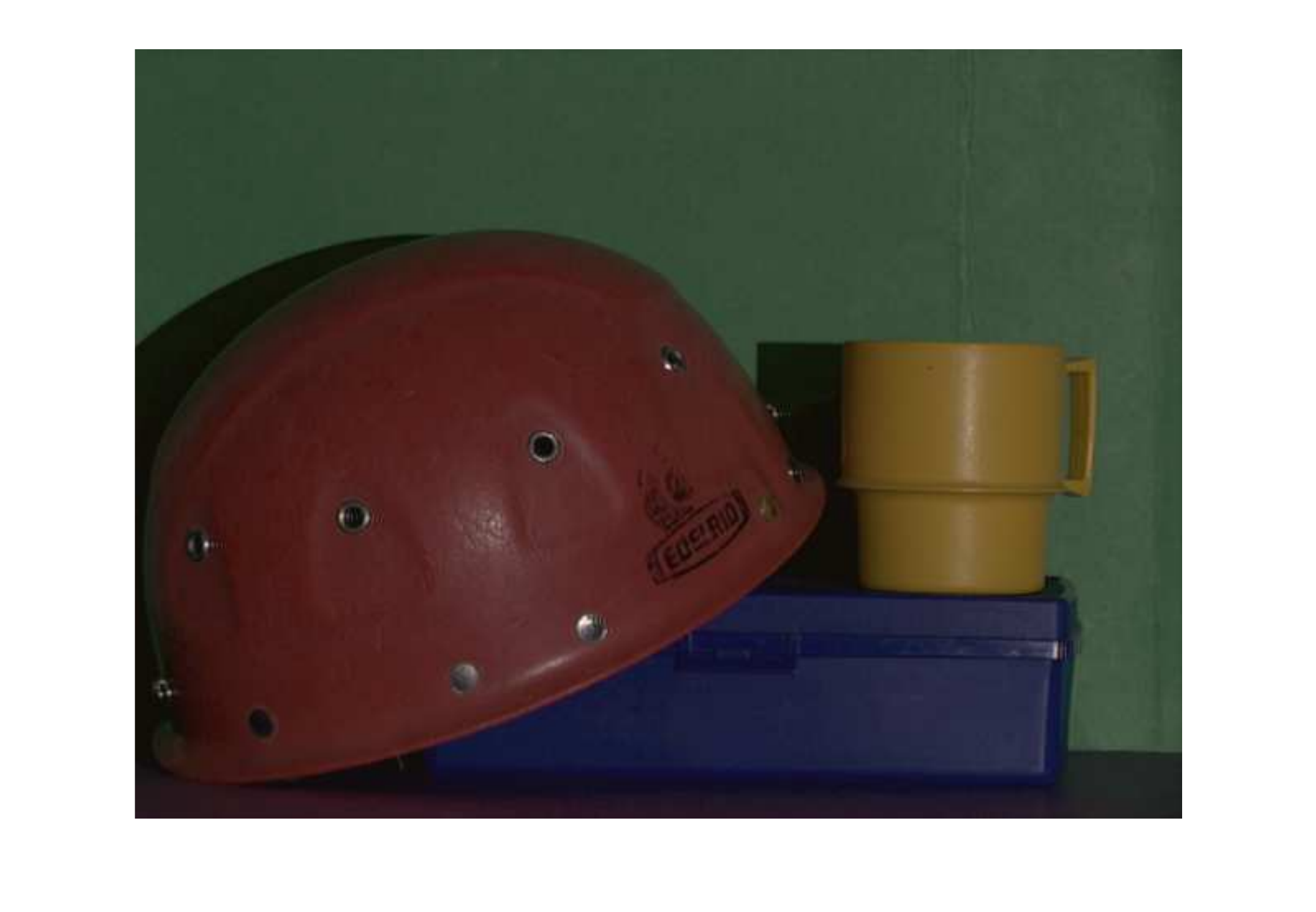}\ }
			
			\\
			
			\subfigure[]{\includegraphics[width=3.75cm]{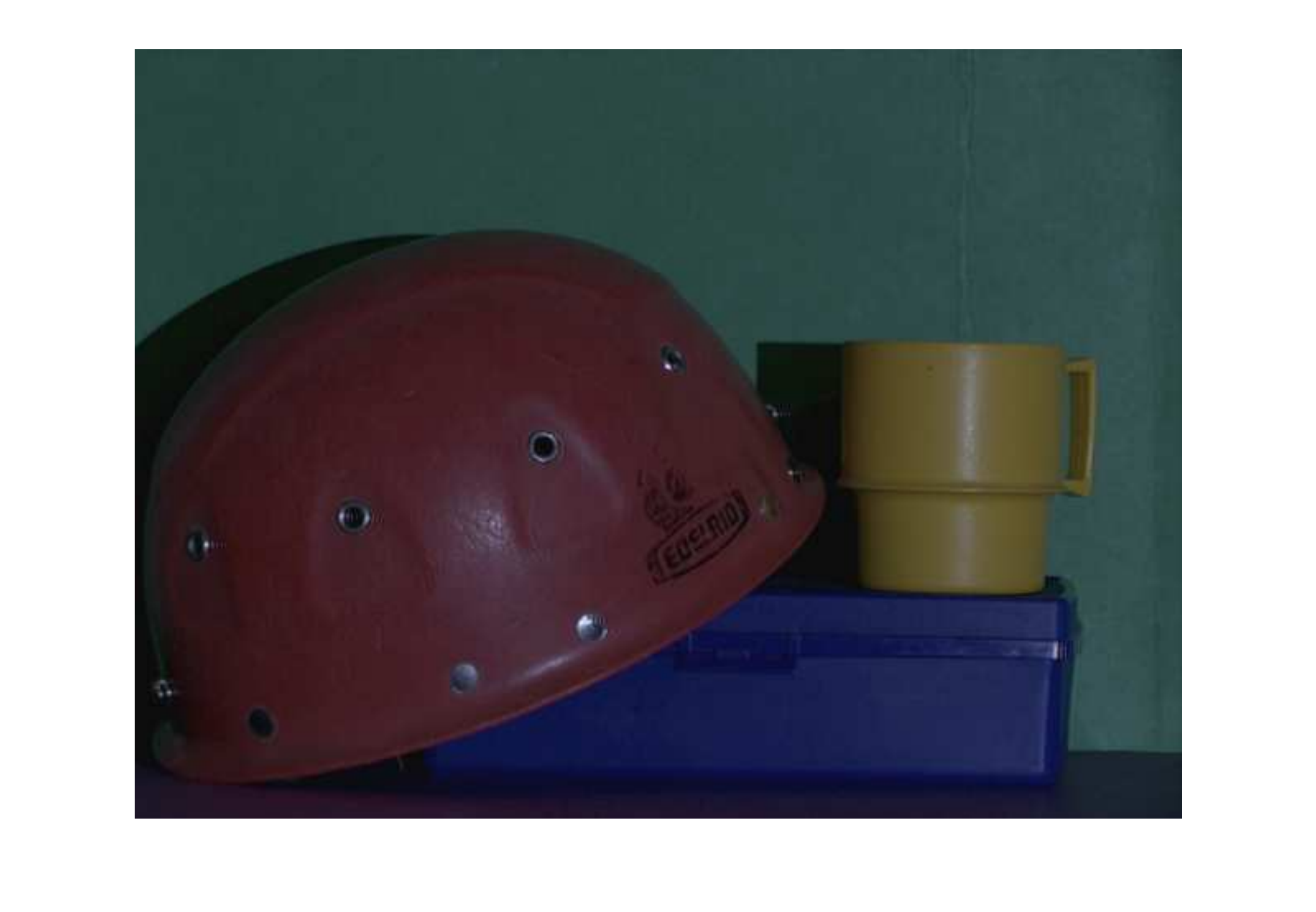}\ }
			\subfigure[]{\includegraphics[width=3.75cm]{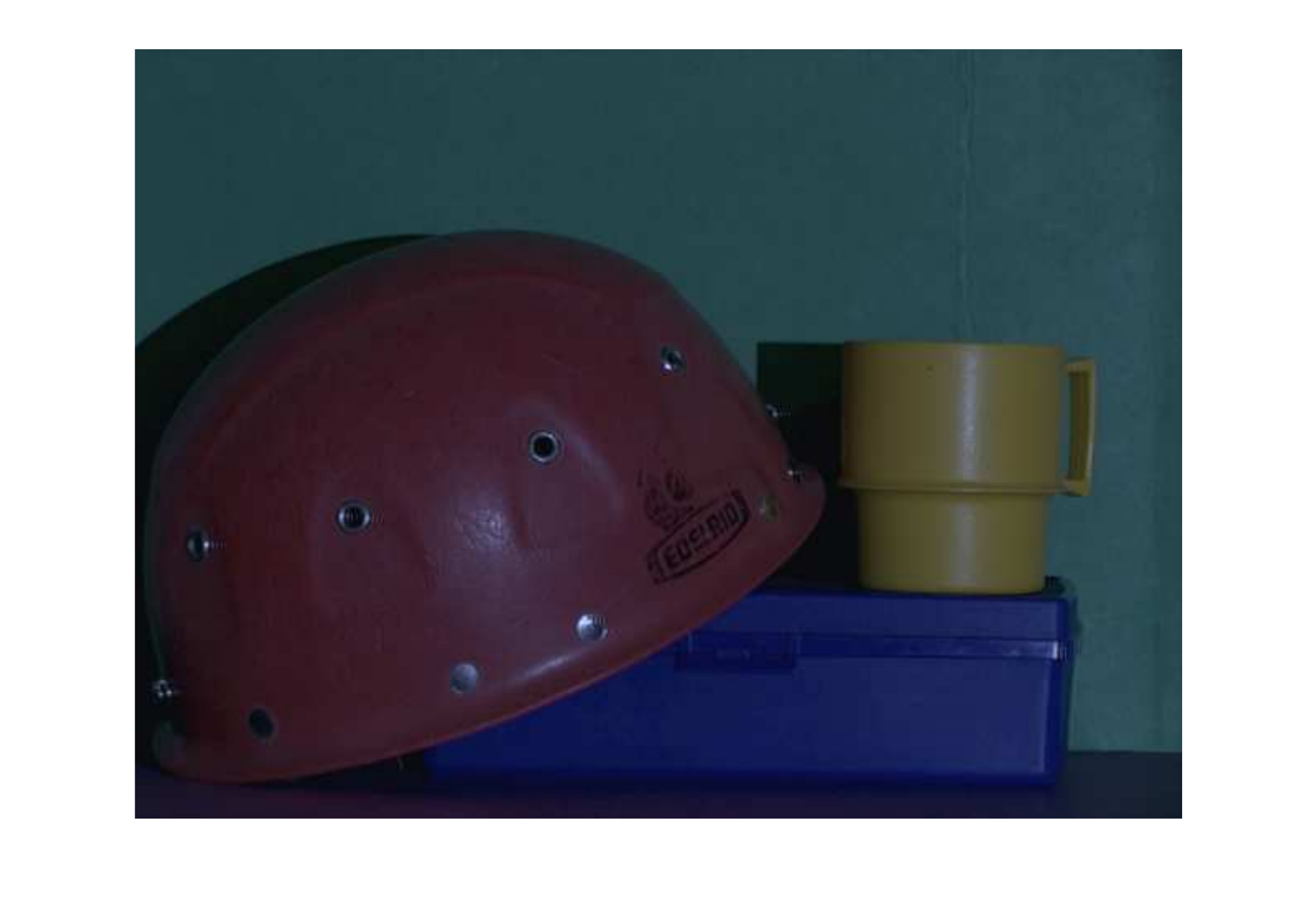}\ }
			\subfigure[]{\includegraphics[width=3.75cm]{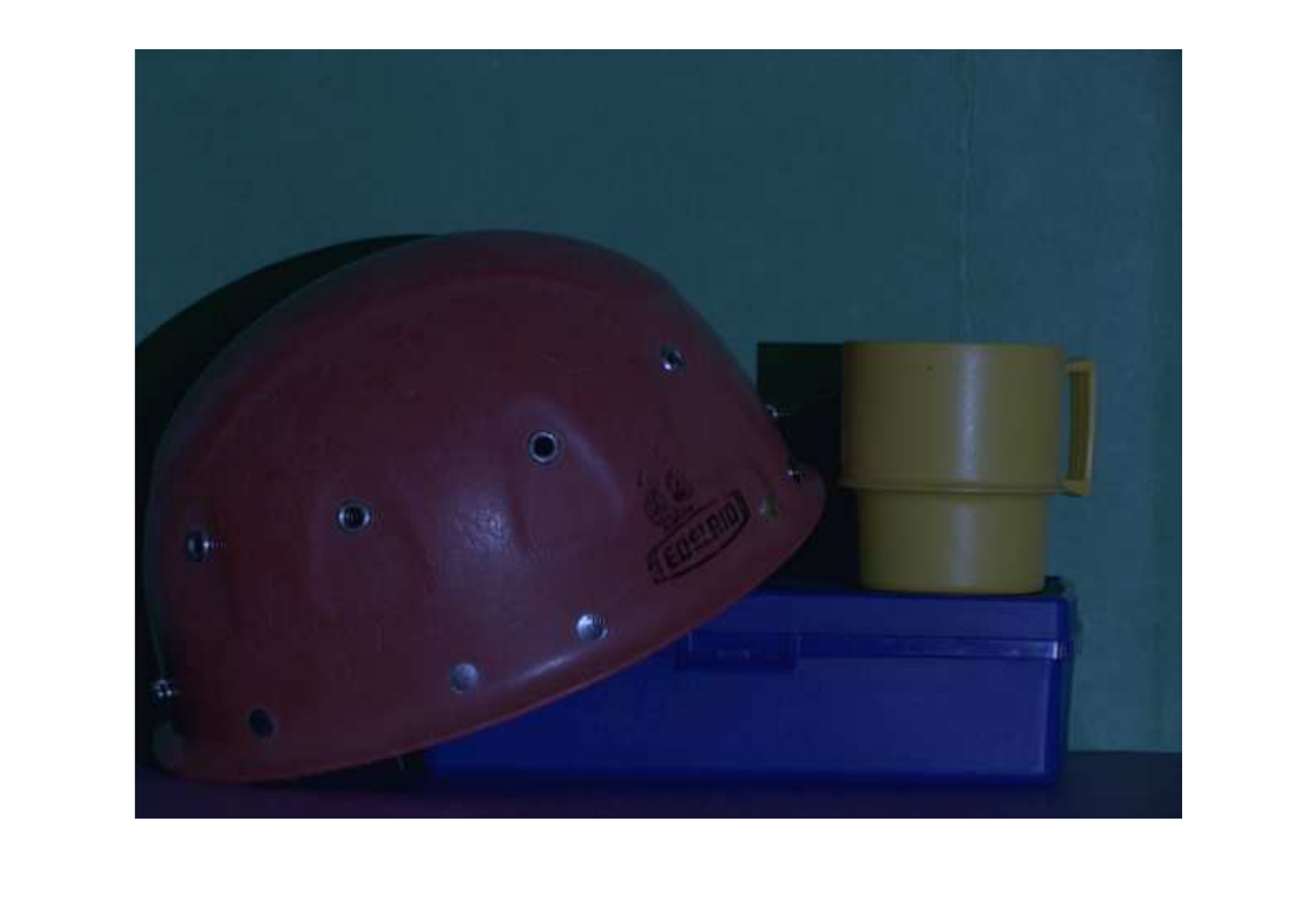}\ }
			\subfigure[]{\includegraphics[width=3.75cm]{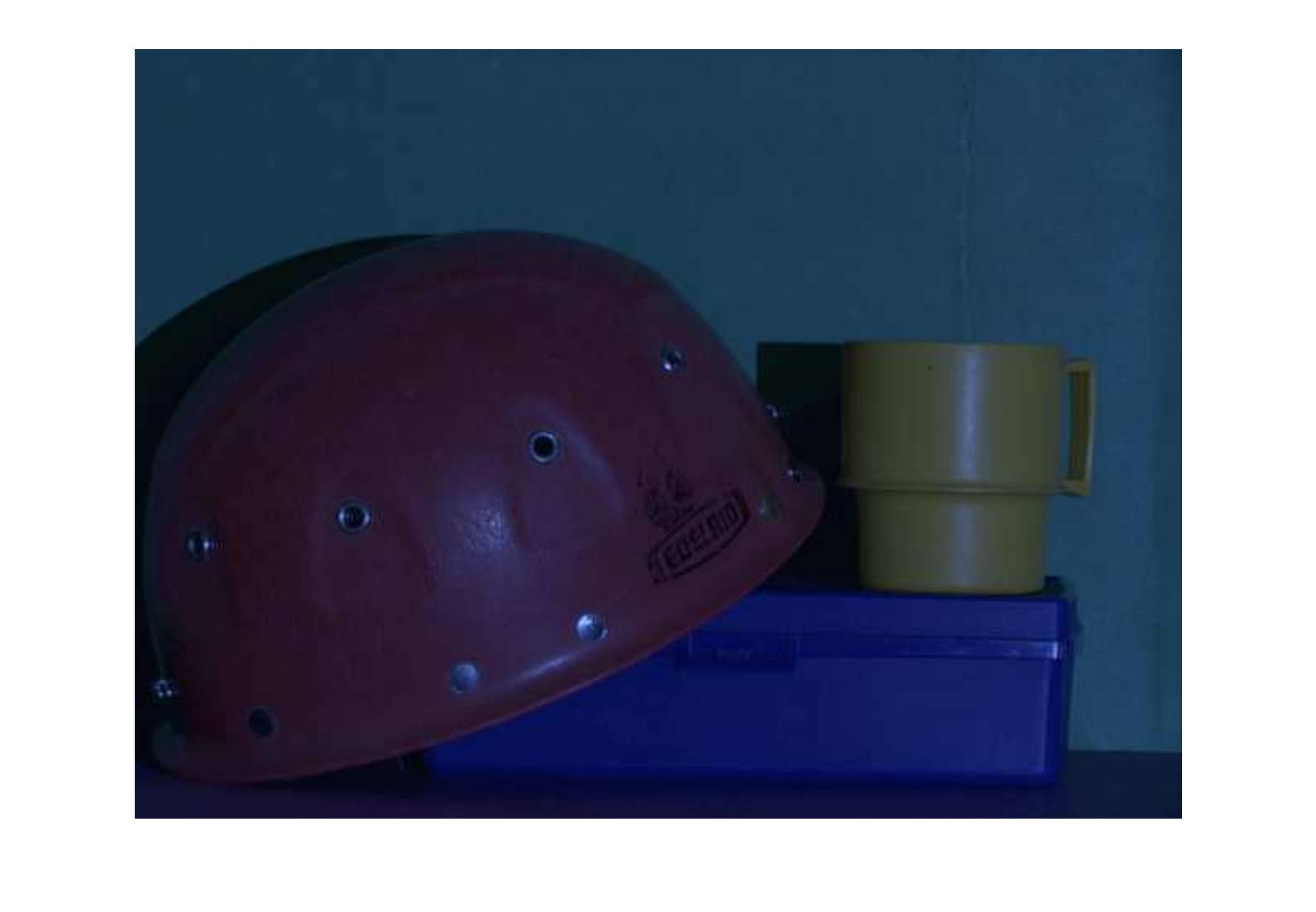}\ }
			
		\end{tabular}
		\vspace*{-0.25in}
		\caption{
			(a-h):\ Images generated by re-lighting with Planckians of differing
			tempertures $T$=1600$^\circ$K, 1900$^\circ$K, 2400$^\circ$K, 2750$^\circ$K, 3900$^\circ$K, 4950$^\circ$K, 6750$^\circ$K, 10600$^\circ$K.
		}
		\label{FIG:RELIGHTSERI}
	\end{center}
	\BUT
\end{figure*}

Fig.~\ref{FIG:RELIGHTSERI} shows the same image for different Planckian
illuminants from 1500$^\circ$K to 10000$^\circ$K, using the proposed re-lighting method.
The method arguably produces reasonable output images corresponding to the
colour of the lights involved.

\begin{figure*}[htbp]
	\begin{center}
		\begin{tabular}[t]{c}
			\subfigure[]{\includegraphics[width=3.75cm]{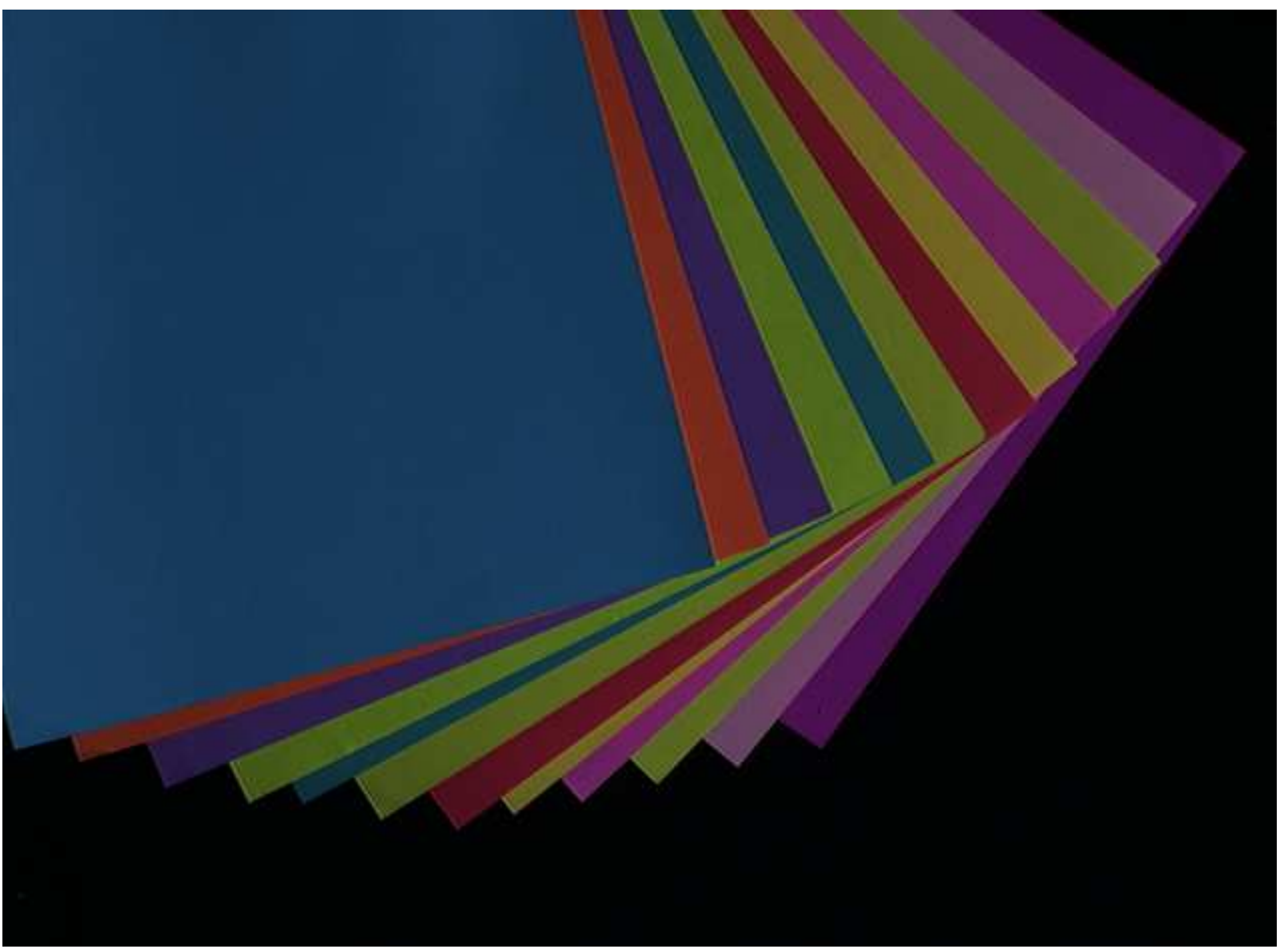}\ }
			\subfigure[]{\includegraphics[width=3.75cm]{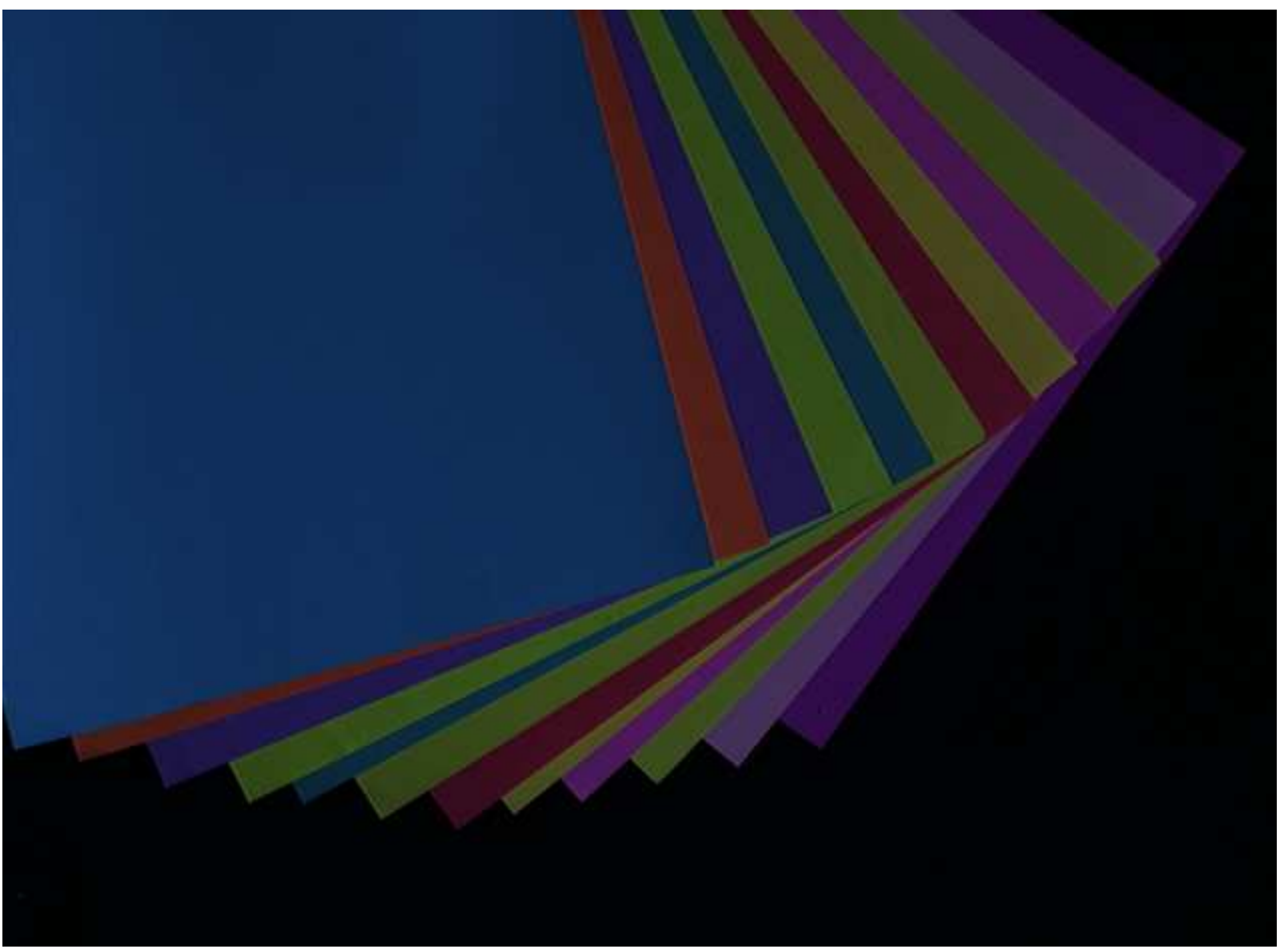}\ }
			\subfigure[]{\includegraphics[width=3.75cm]{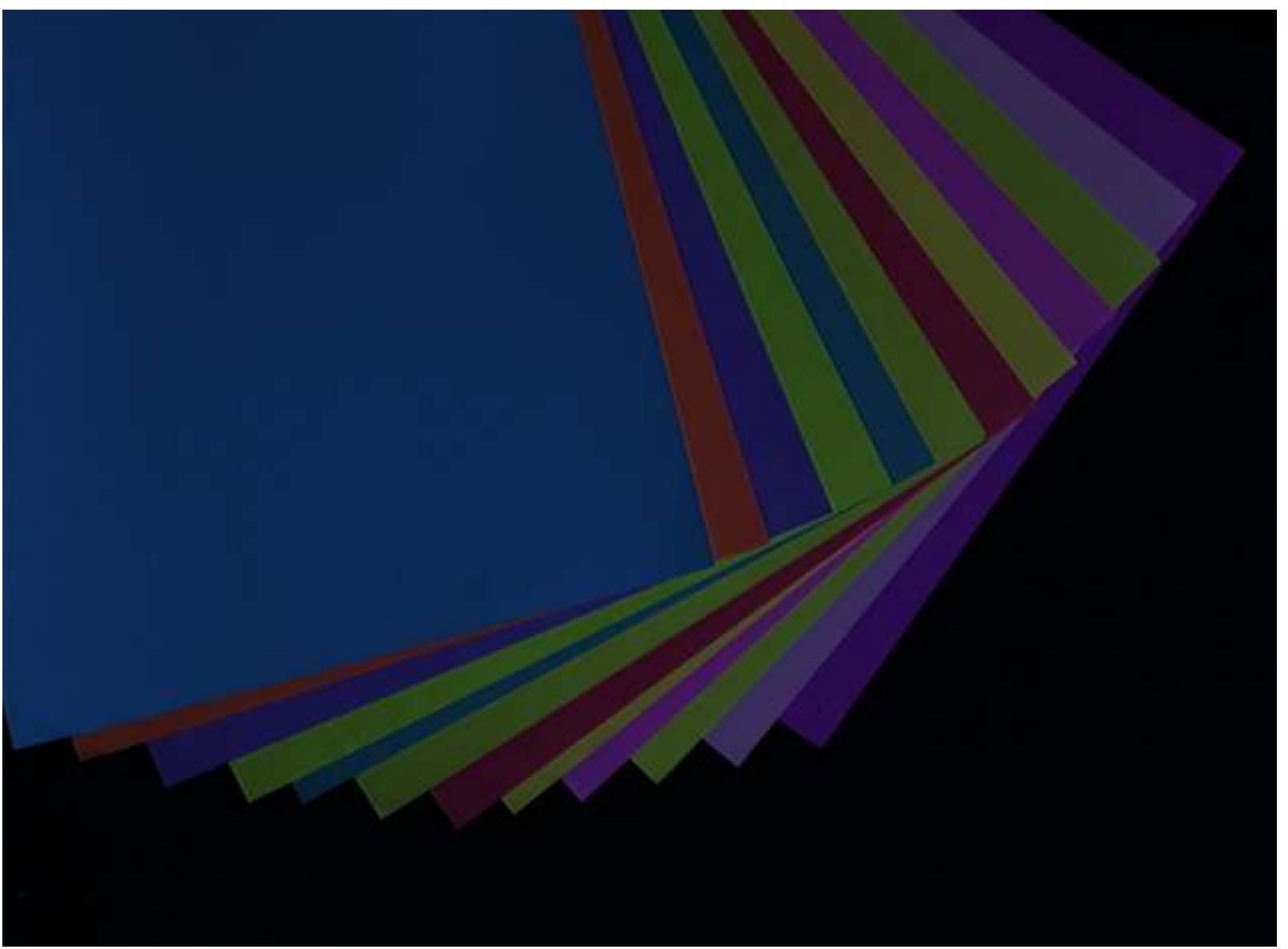}\ }
			\subfigure[]{\includegraphics[width=3.75cm]{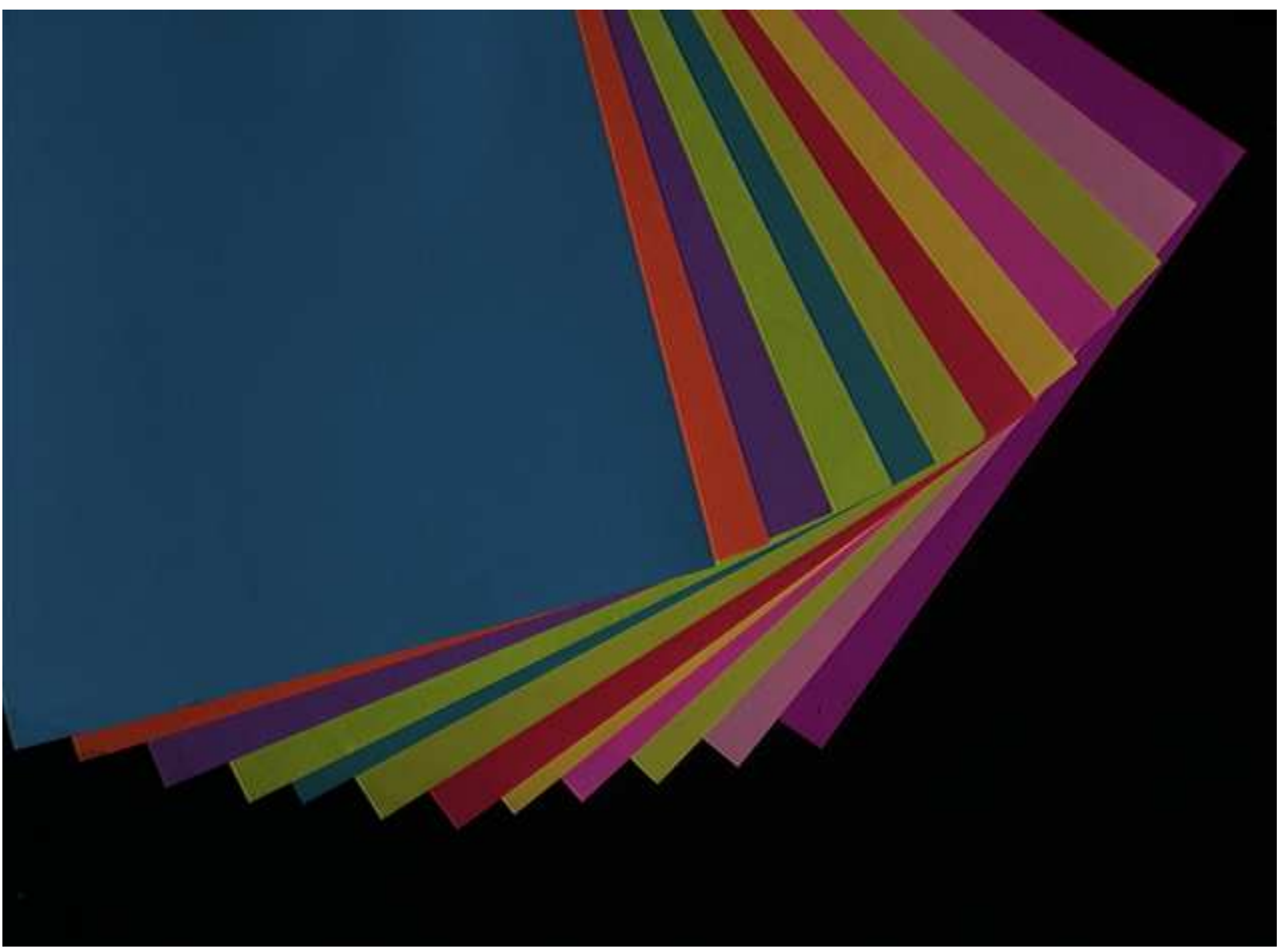}\ }
			
			\\
			
			\hspace{3.85cm}
			\subfigure[]{\includegraphics[width=3.75cm]{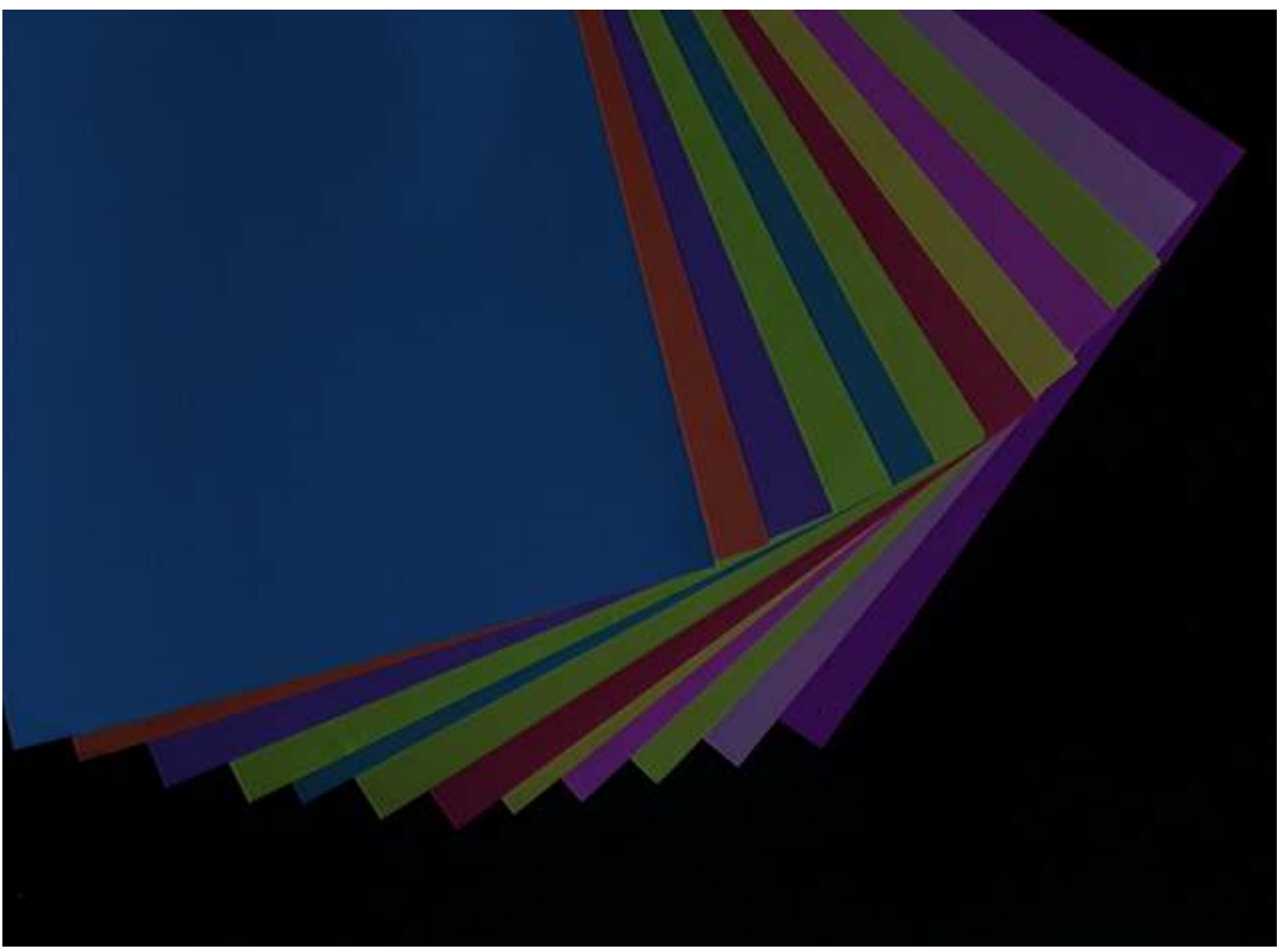}\
			}
			\subfigure[]{\includegraphics[width=3.75cm]{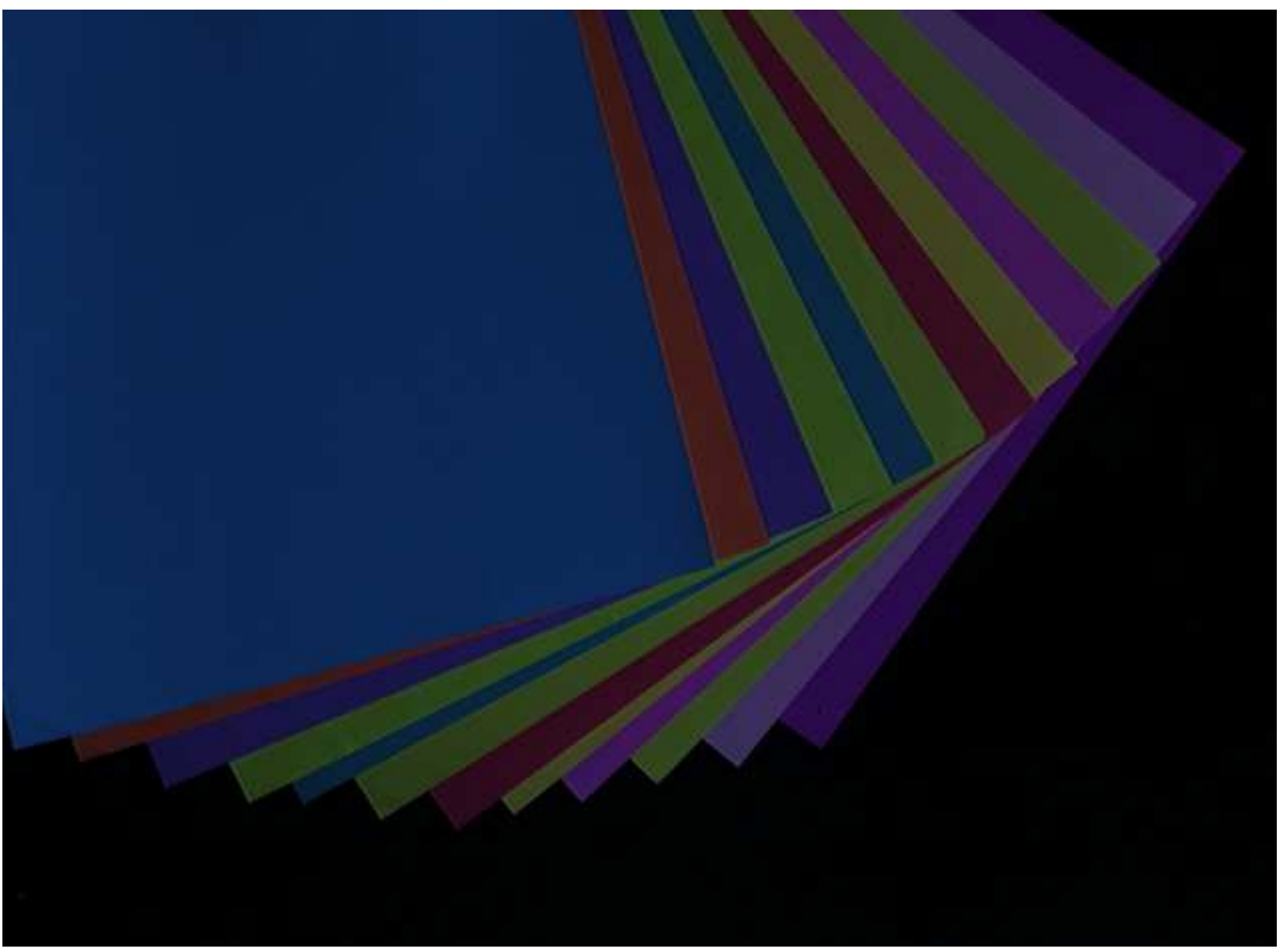}\
			}
			\subfigure[]{\includegraphics[width=3.75cm]{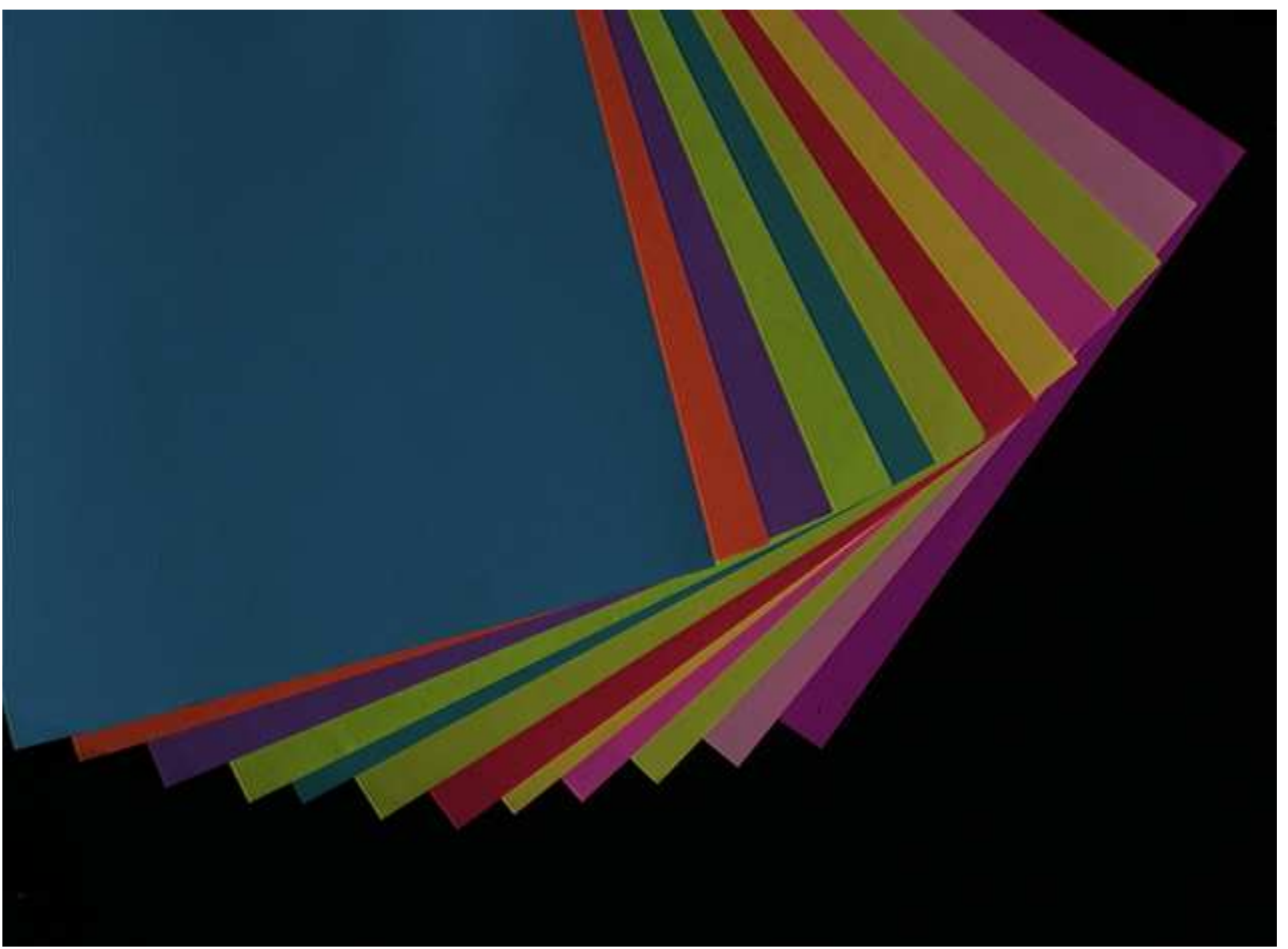}\
			}
			
		\end{tabular}
		\vspace*{-0.25in}
		\caption{
			(a-d):\ Input image; (e-g):\ Images generated by re-lighting of images (a) using
			(b-d) estimated illuminants on daylight locus. The PSNR for (e-g) are
			respectively: $42.3$, $37.1$ and $33.9$.}
		\label{FIG:RELIGHTSERI2}
	\end{center}
	\BUT
\end{figure*}

In another experiment, we compare the error of using daylight locus for
re-lighting, via eq.~(\ref{eq:diagTransform})
compared to using 
the actual value of illuminants. 
Fig.~\ref{FIG:RELIGHTSERI2} shows the same image transferred to other measured
images, using their estimated illuminants on the daylight locus. 
In all, we generated re-lit images 
for a fixed object under 8 different illuminants (56
re-lightings). In terms of PNSR error for generated images compared to measured
ones,
we found a median PSNR value of $33.8$dB, with minimum and maximum values of
$28.2$ and $43.5$dB. These values demonstrate acceptable faithfulness of
rendition for images under new lighting.
QAs another comparison, instead of using illuminants on the locus we instead
used actual measured illuminants in transforming the image via
eq.~(\ref{eq:diagTransform}). Now the min/median/max PSNR values are $28.2$,
$34.0$ and $43.2$, almost identical with those found using the illuminant
approximation derived from the locus. This demonstrates that using the locus is
nearly as good as using the actual illuminant, for this re-lighting task, with
negligible difference in results.

\section{Matte Image from Angle Image \label{SEC:MATTE}}

We would like to generate a matte output image, which will then act as an
invariant image free of shading and specularity (which could then be used as
input to a segmentation scheme, for example).  However, our specular-invariant quantity is the
angle from the recovered specular point, to each image pixel in feature space.
However, this angle
encapsulates hue information.
The main point is that 
the angle from the specular point to the feature point of a
pixel is approximately independent of 
the presence or absence of specular content at that pixel. 
Hence, if there is any structure in the image feature space \mbchi
from specular content, then by going over to this 2-D chromaticity space radii
from the specular point will be in the same direction for pixels of the same
body colour with or without specular content.

Based on the chromaticity-space model \cite{LEE.HC.ILLUM.CHROM.86}, a pixel value is
a linear combination of the light colour and the matte colour, as measured by
the camera, resulting in a line in chromaticity space starting from the matte
point for any particular colour and leading towards the illuminant colour. 
Since 
we already know the light, assumed to be the colour of the specular point, 
we have this line direction for each pixel, leading from from specular point
to that pixel.
Moreover,
these lines correspond to the angular values that we already assigned to each pixel. 
We can therefore consider the pixels with the same angular value as belonging to
the same matte object  --- although in real images it is possible that two matte
values fall on the same line toward the specular point. Here we initially simply take any
such cases as belonging to the same matte value;  however, below, considering spatial information
we can in fact separate these two matte values from each other.

To make the calculation simpler we transform the chromaticity of the specular
point to the origin and use polar coordinate $(r,\theta)$. We discretize angle
values by using $\left\lfloor \theta \right\rfloor$ to have 360 bins. Therefore
for each chromaticity point $\mbv$, we consider $(r_v,\theta_v)=polar(v-S)$, where $S$ is the specular point.

\begin{figure}[htbp]
	\begin{center}
		\begin{tabular}[t]{c}
			\subfigure[]{\includegraphics[width=4.5cm]{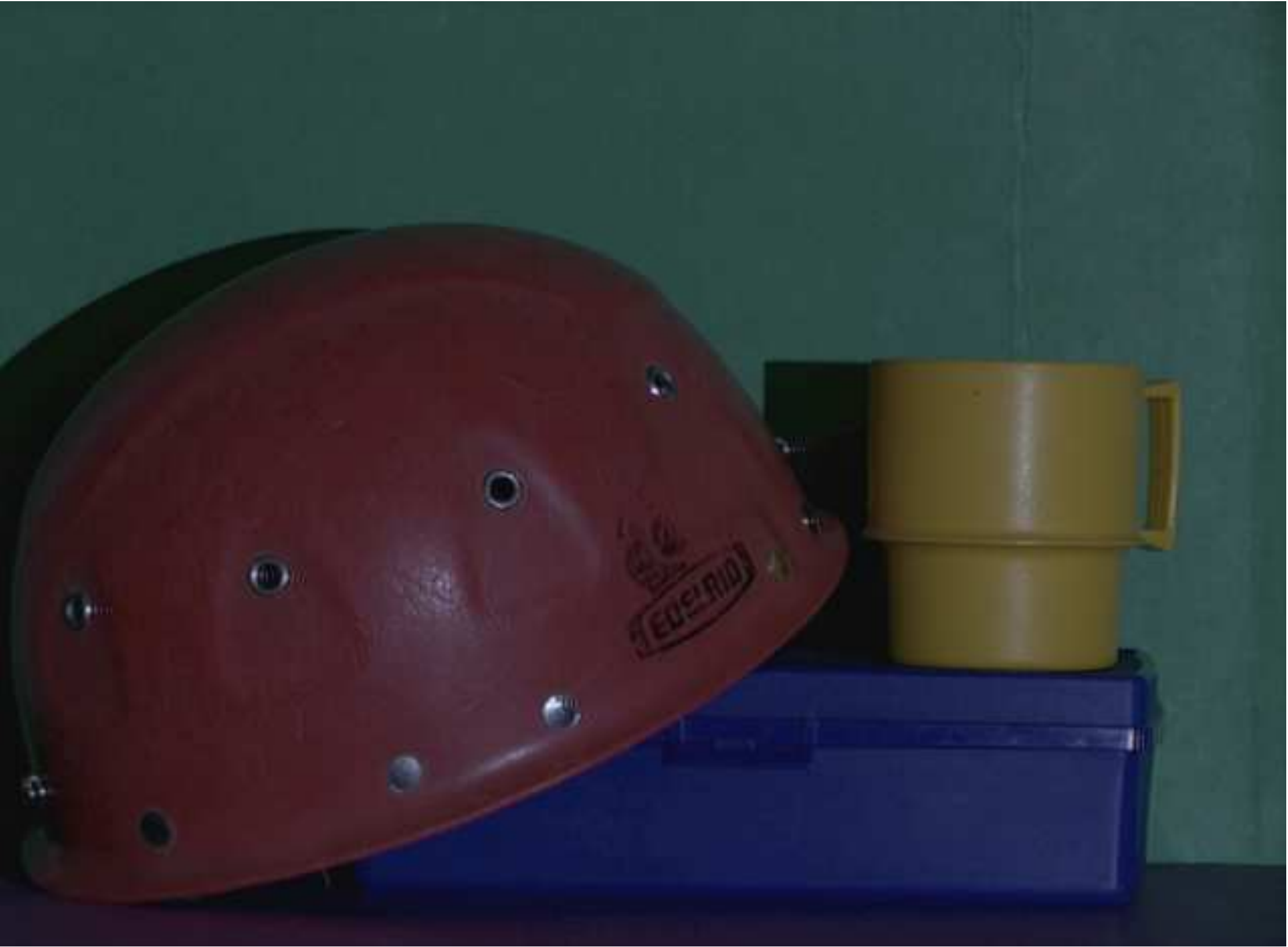}\ }
			\subfigure[]{\includegraphics[width=5.5cm]{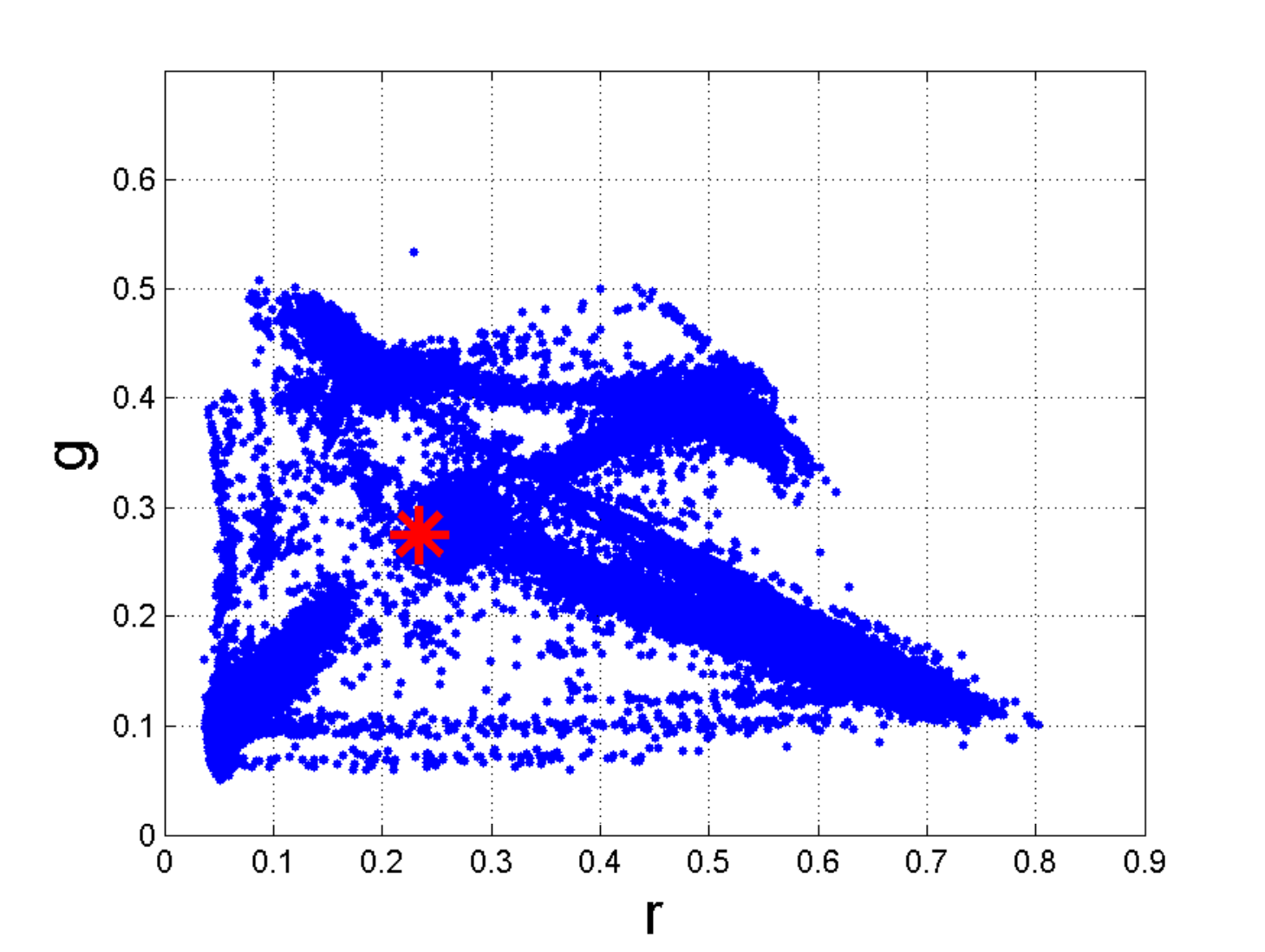}\ }
			\subfigure[]{\includegraphics[width=5.5cm]{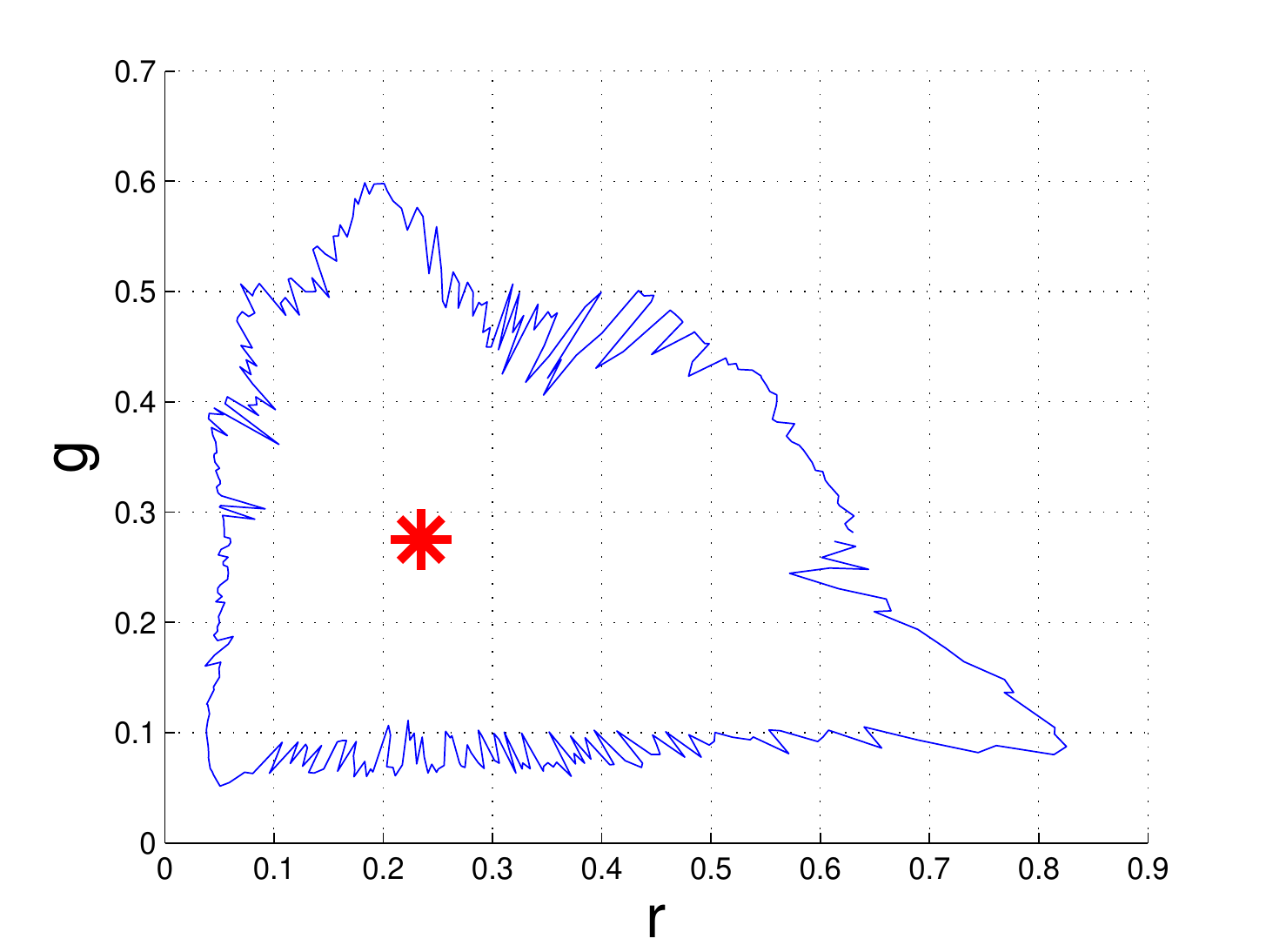}\ }
		\end{tabular}
		\vspace*{-0.25in}
		\caption{
			(a):\ A real image.
			(b):\ Chromaticity points for (a); the red star is the correct specular point.
			(d):\ Angular projection to matte colour for image points.
		}
		\label{FIG:PROJECT}
	\end{center}
	\BUT
\end{figure}

The final step to generate a specularity-free colour image is to find a matte value for each pixel. 
We take the farthest-most pixel from the specular point (i.e., maximum radius $r$) for each $\theta$ as 
the matte colour (after removing outliers). 
So the matte colour for each pixel at that angle is identified with the farthest pixel. 
We call this process ``angular projection to matte colour". 
In other words we are projecting chromaticity points to the border of
chromaticity values for each angle, considering the specular point as the center
of projection:
\begin{equation}
matte(\mbv)=maxIndex_{\theta_u=\theta_v}(r_u)
\label{eq:farest}
\end{equation}
Fig.~\ref{FIG:PROJECT} illustrates the projection for chromaticity points for a real image by angular projection to matte colour.

The angular projection is more sensitive to noise the closer are image feature
points to the specular point.
Generally,
because of noise angular projection to matte colour may completely fail for
highlights. Hence we deal with the 10\% of pixels that are close to each candidate specular point differently
--- we iteratively inpaint these pixels using matte colour data from
neighbouring pixels
that correspond to the same angular value (1-D  inpainting).
That is, we use
voting based on the matte colour of the pixel's neighbours:
the new matte colour for that pixel will be the majority of its neighbours' matte
colour if it garners at least half of the votes.

\begin{figure}[htbp]
	\begin{center}
		\begin{tabular}[t]{c}
			\subfigure[]{\includegraphics[width=5.25cm]{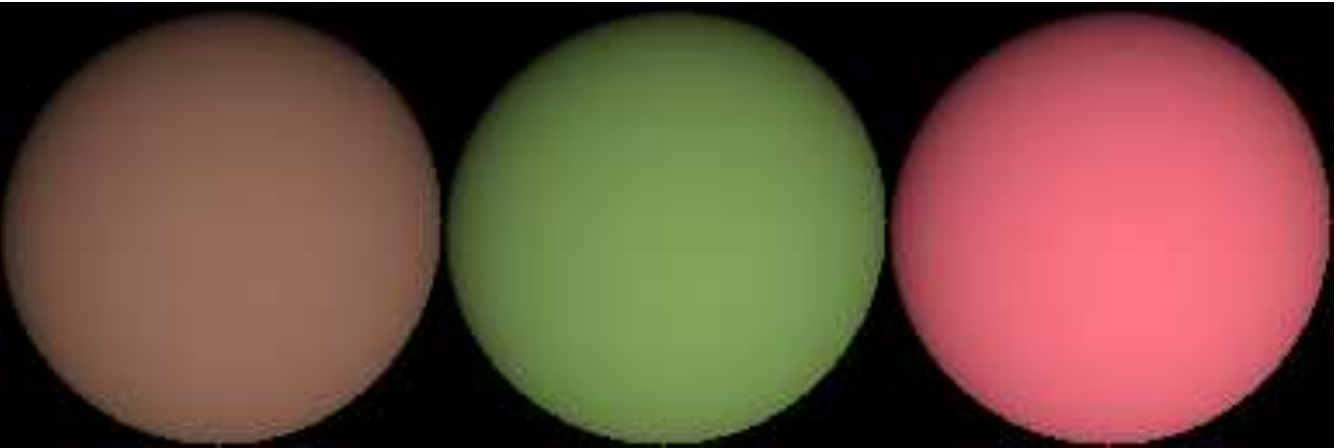} } 
			\subfigure[]{\includegraphics[width=5.25cm]{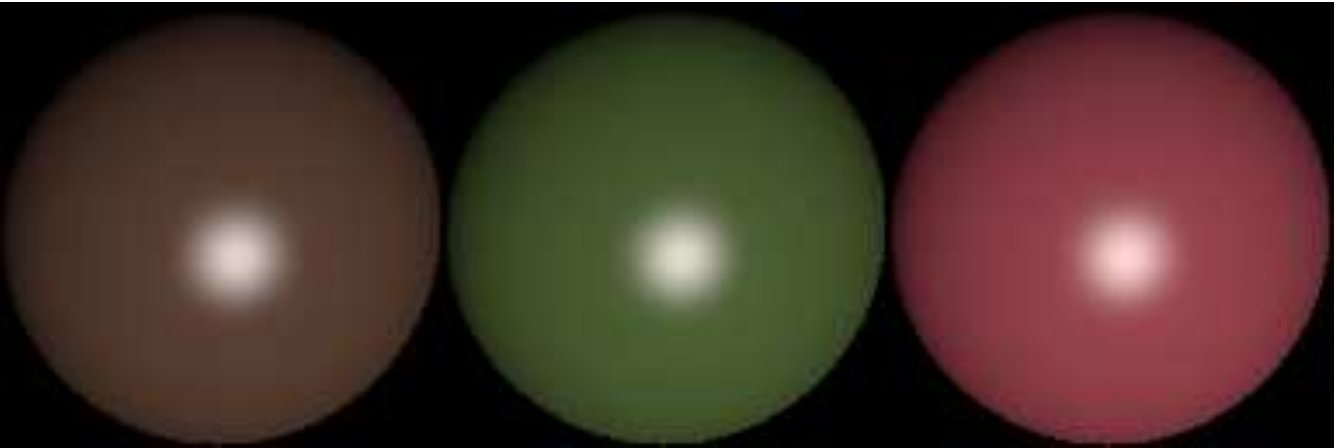} }
			\\
			\subfigure[]{\includegraphics[width=5.25cm]{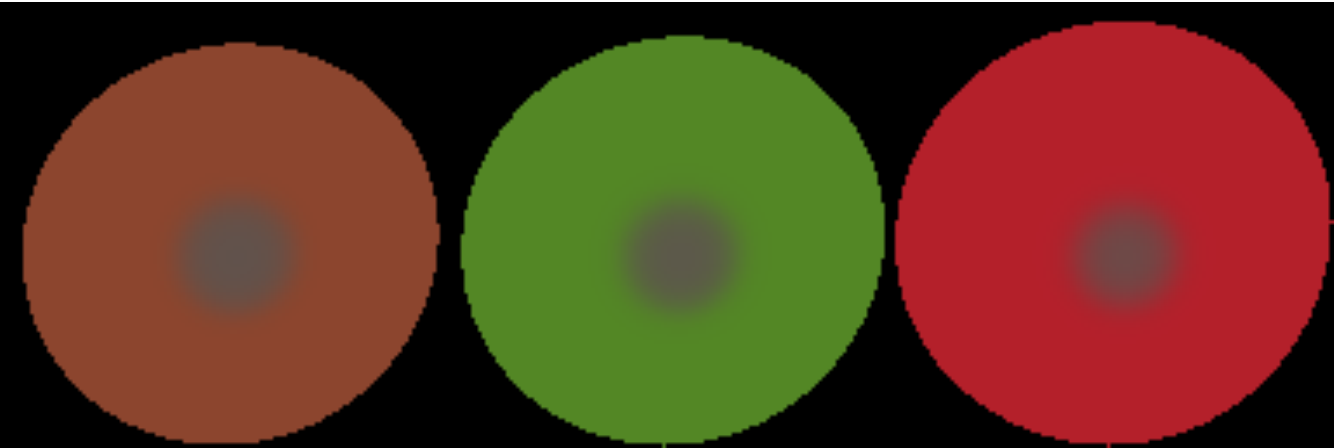}\ }
	     	\subfigure[]{\includegraphics[width=5.25cm]{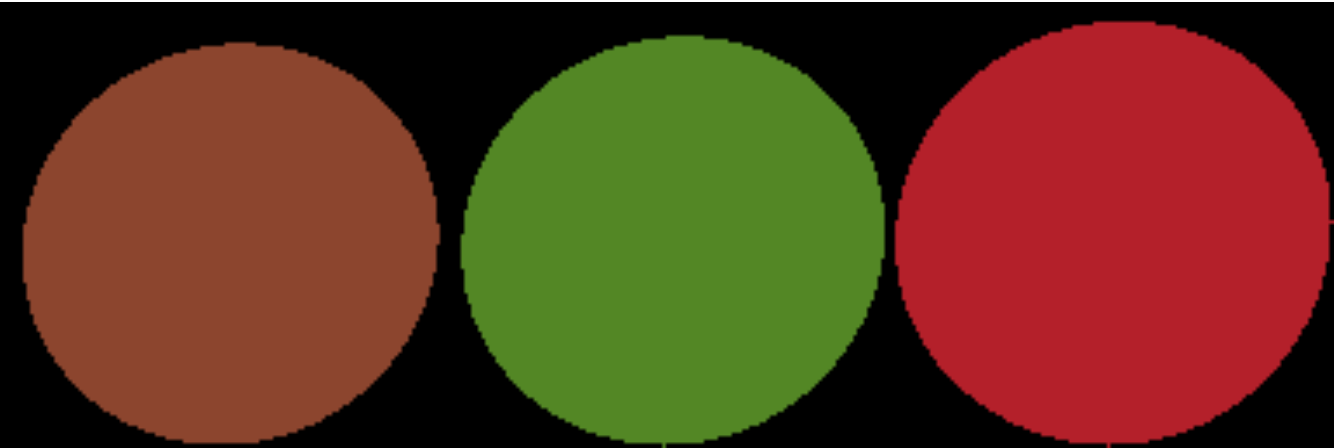}\ }
	    \end{tabular}
		\vspace*{-0.25in}
		\caption{
			(a):\ Ground-truth synthetic image of Lambertian surfaces under standard illuminant D65.
			(b):\ Image with specular reflection added.
			(c):\ The {\em input(b)} chromaticity image.
			(c):\ The chromaticity image resulting from angular projection to matte colour.
			}
		\label{FIG:SYNRESULT}
	\end{center}
	\BUT
\end{figure}

A complete synthetic example consists of accurately modeled matte plus
specular components. Here, let us consider
a test image consisting of three shaded spheres (as in
\cite{FINLAYSON.DREW.ICCV01}), with
surface colours equal to patches 1, 4, and 9
of the Macbeth ColourChecker \cite{MCCAMY.MACBETH.76} \index{Macbeth ColourChecker}
(dark skin, moderate olive green, moderate red), and under
standard illuminant D65 \index{D65} (standard daylight with correlated colour
temperature 6500K \cite{WYSZECKI82}) using the sensor curves for a Kodak DCS420
digital colour camera. If we adopt a Lambertian model
then the matte image is as in 
Fig.~\ref{FIG:SYNRESULT}(a).
We now add 
a specular reflectance lobe for each surface reflectance function.
We use the Phong illumination model\cite{FOLEY.VANDAM.90}, together with
underlying matte Lambertian
shading.
Here, we use a Phong factor of 1 for the  magnitude relative to matte.
For the Phong power, we use a power of 20, where the inverse is basically
roughness, 0.05. The matte image goes over to one with highlights as in
Fig.~\ref{FIG:SYNRESULT}(b).
For our synthetic example, the resulting chromaticity image \mbrho is shown in Fig.~\ref{FIG:SYNRESULT}(d). Comparing to the {\em
	input} chromaticity image in Fig.~\ref{FIG:SYNRESULT}(c), we see that the algorithm performs very
well for generating the underlying matte image --- specularities in the center of each sphere are
essentially gone.  In comparison, Fig.~\ref{FIG:SYNRESULT}(a) 
shows the theoretical, correct, matte image, which is indeed very close to the algorithm output
in Fig.~\ref{FIG:SYNRESULT}(a). 

\vspace*{0.1in}

\begin{figure}[htbp]
	\begin{center}
		\begin{tabular}[t]{c}
			
			\includegraphics[width=3.75cm]{plastic-2_solux-4700_tif_tosrgb.pdf}\ 
			\includegraphics[width=4cm]{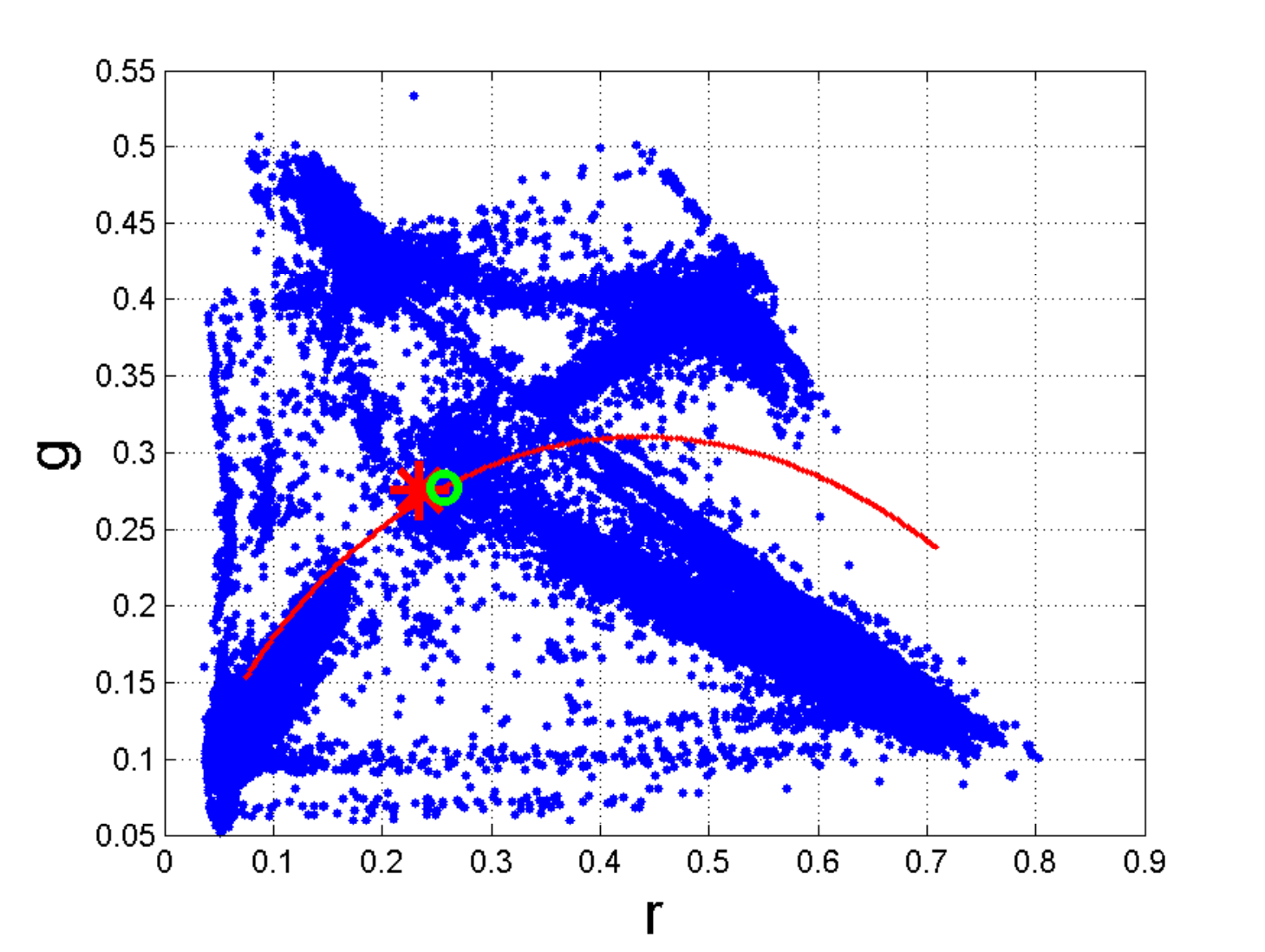}\ 
			\includegraphics[width=3.75cm]{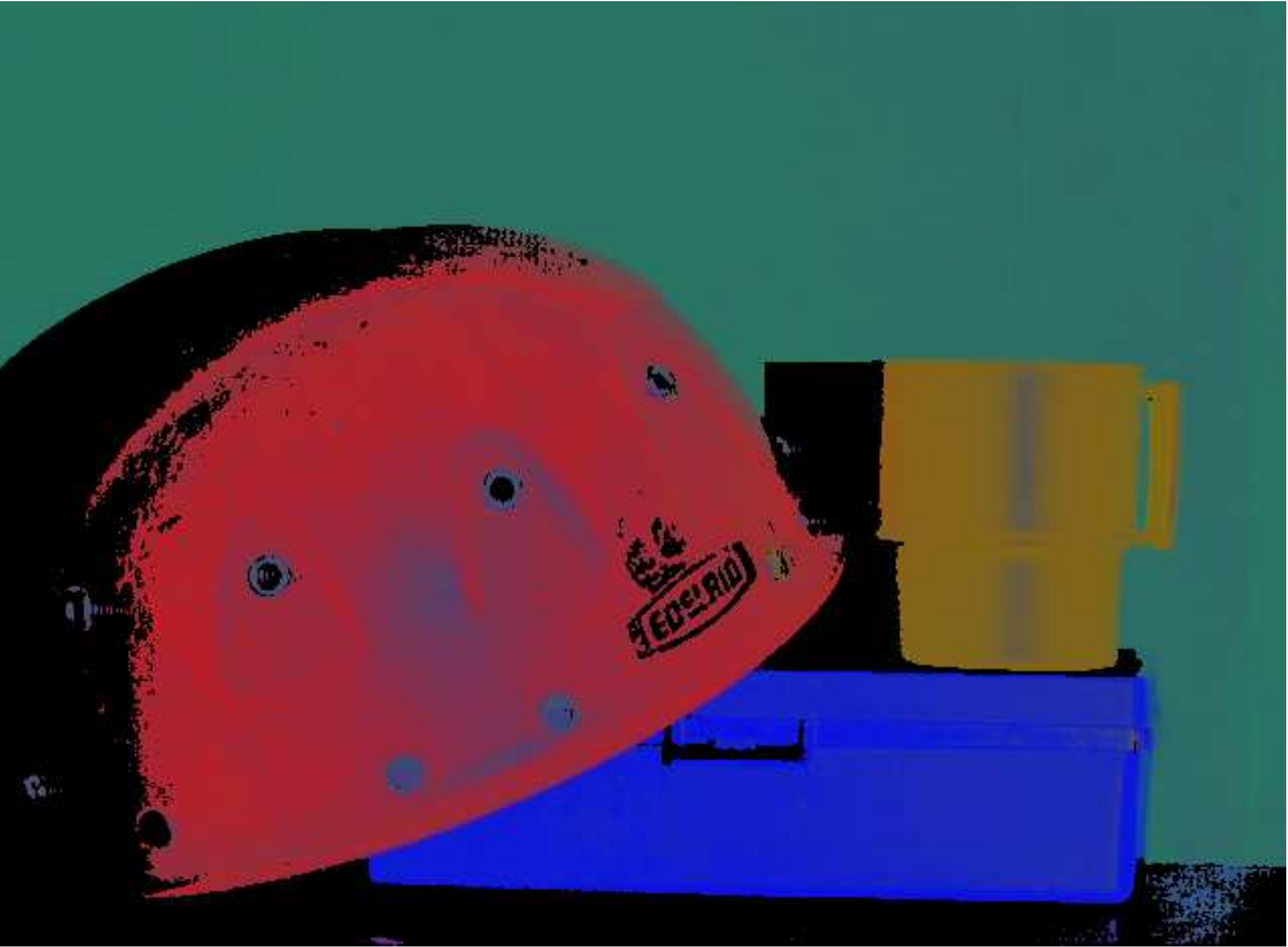}\ 
			\includegraphics[width=3.75cm]{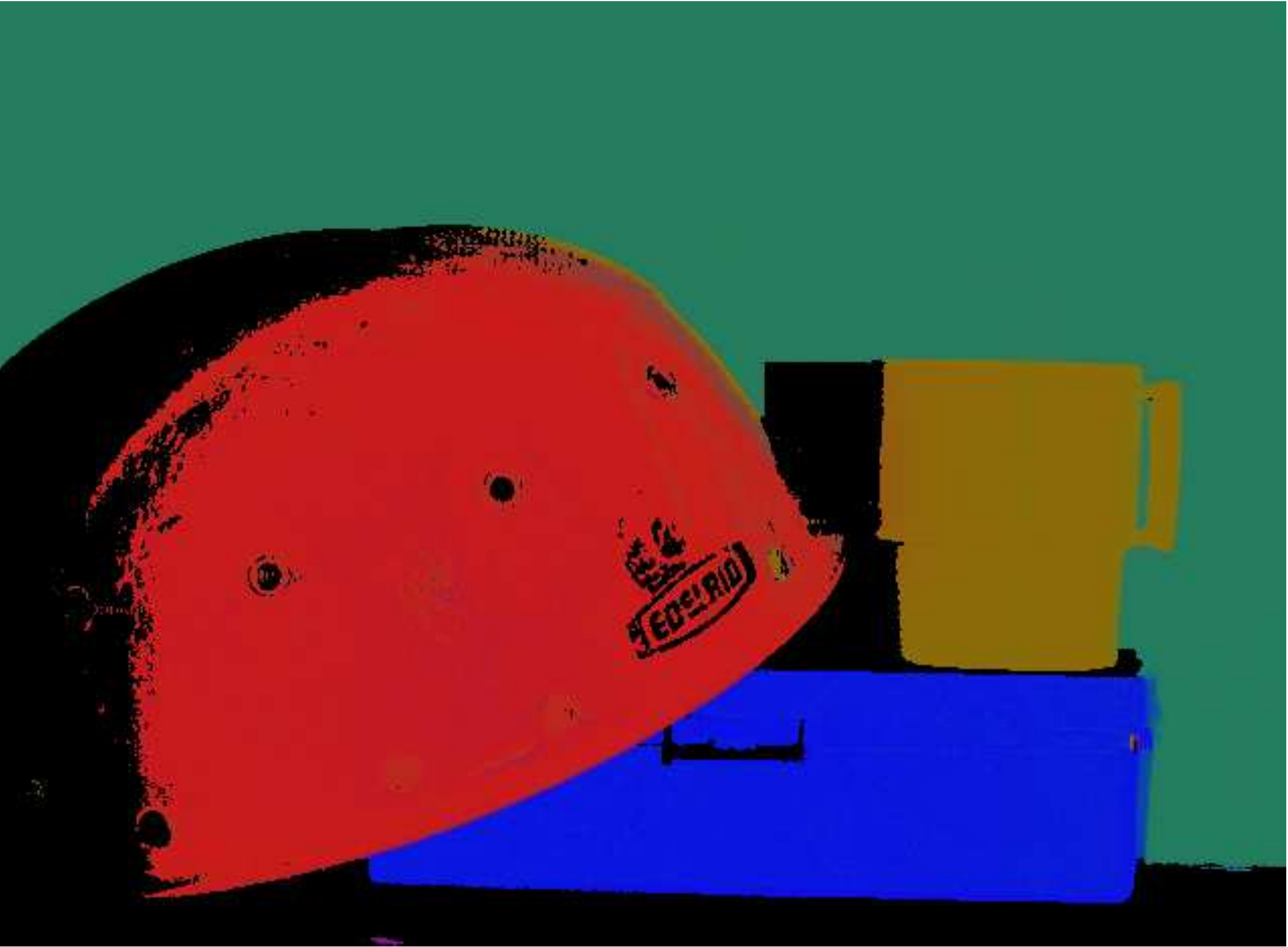}\ 
			
			\\
			
			\includegraphics[width=3.75cm]{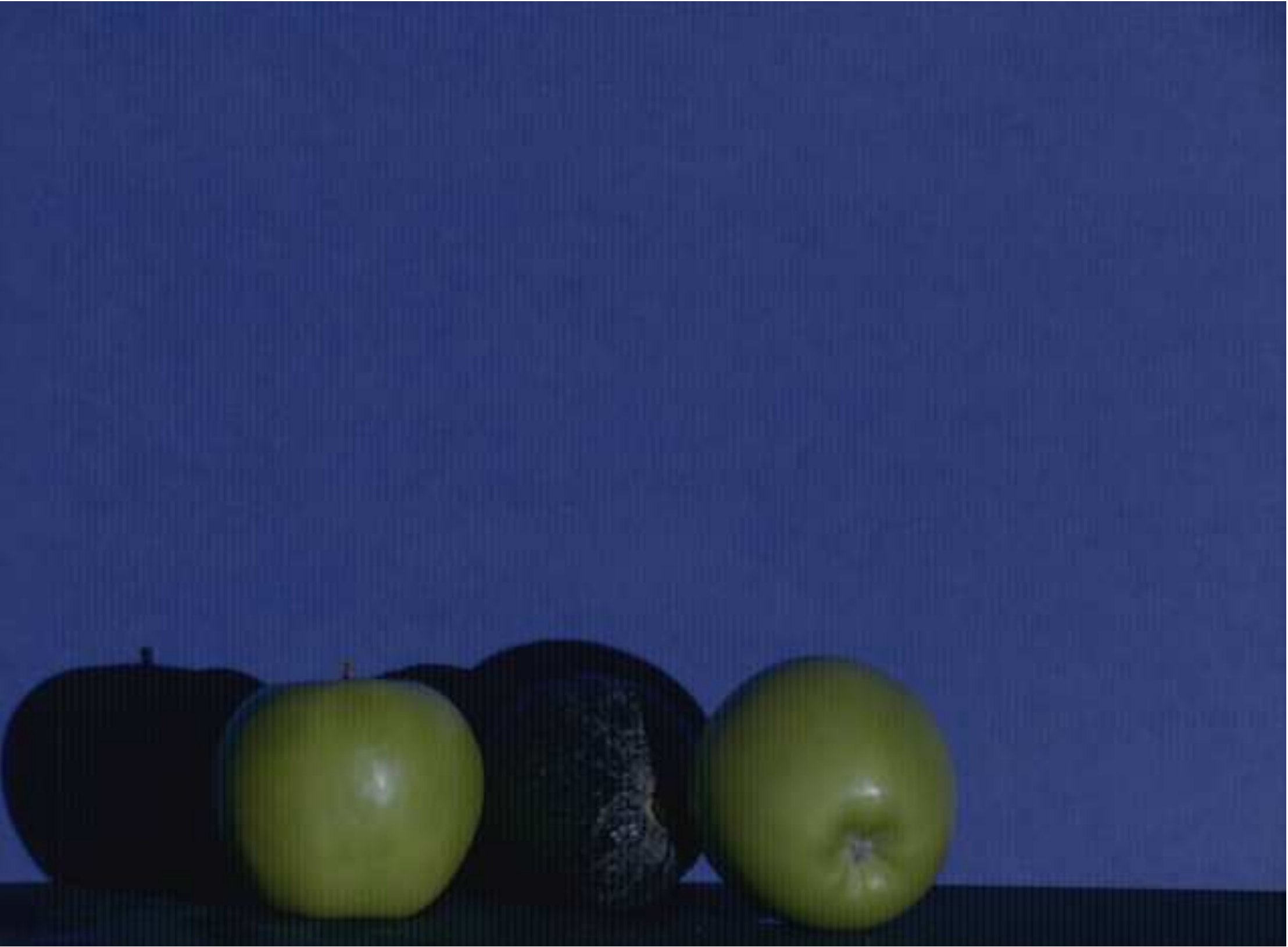}\ 
			\includegraphics[width=4cm]{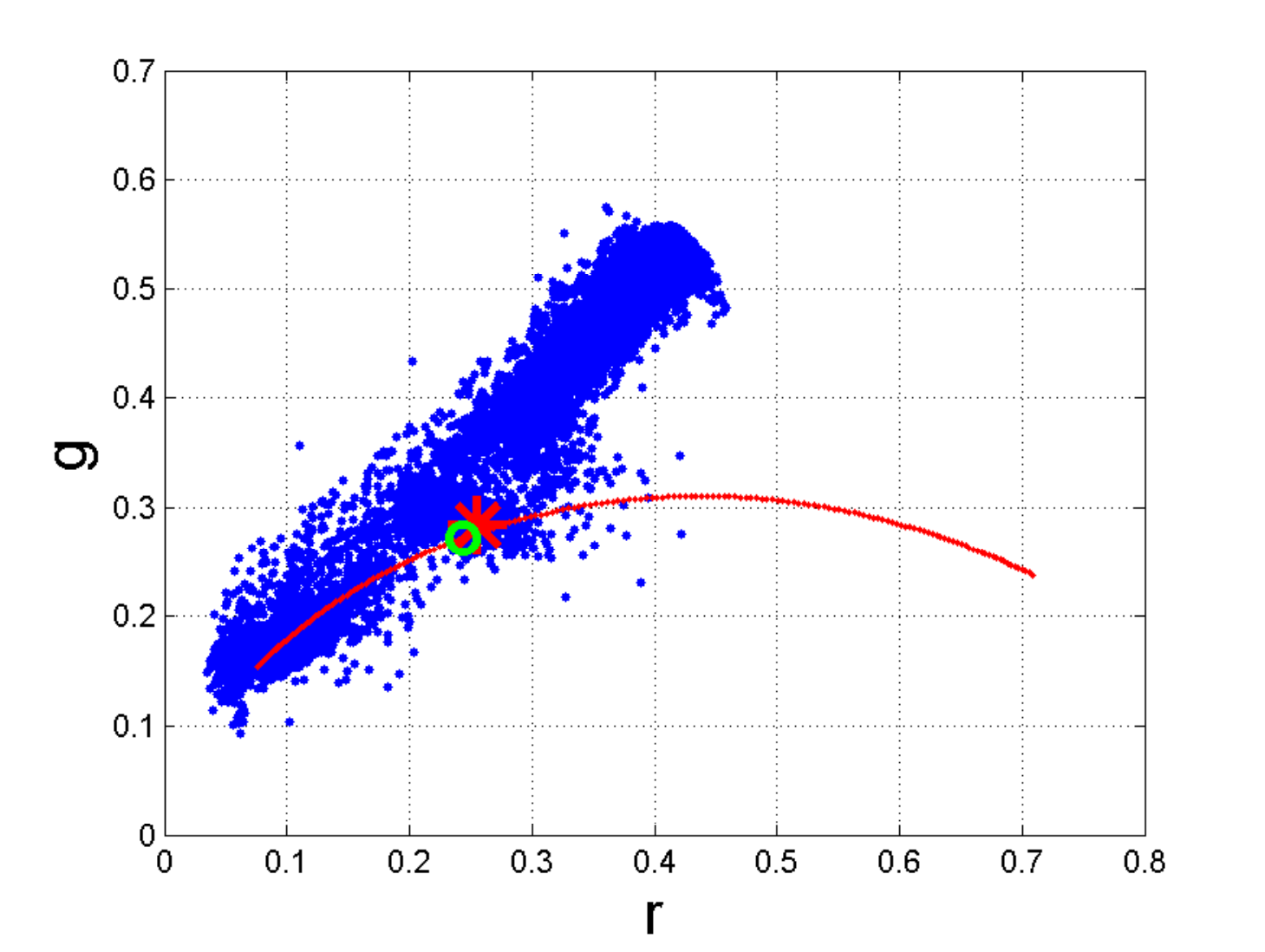}\ 
			\includegraphics[width=3.75cm]{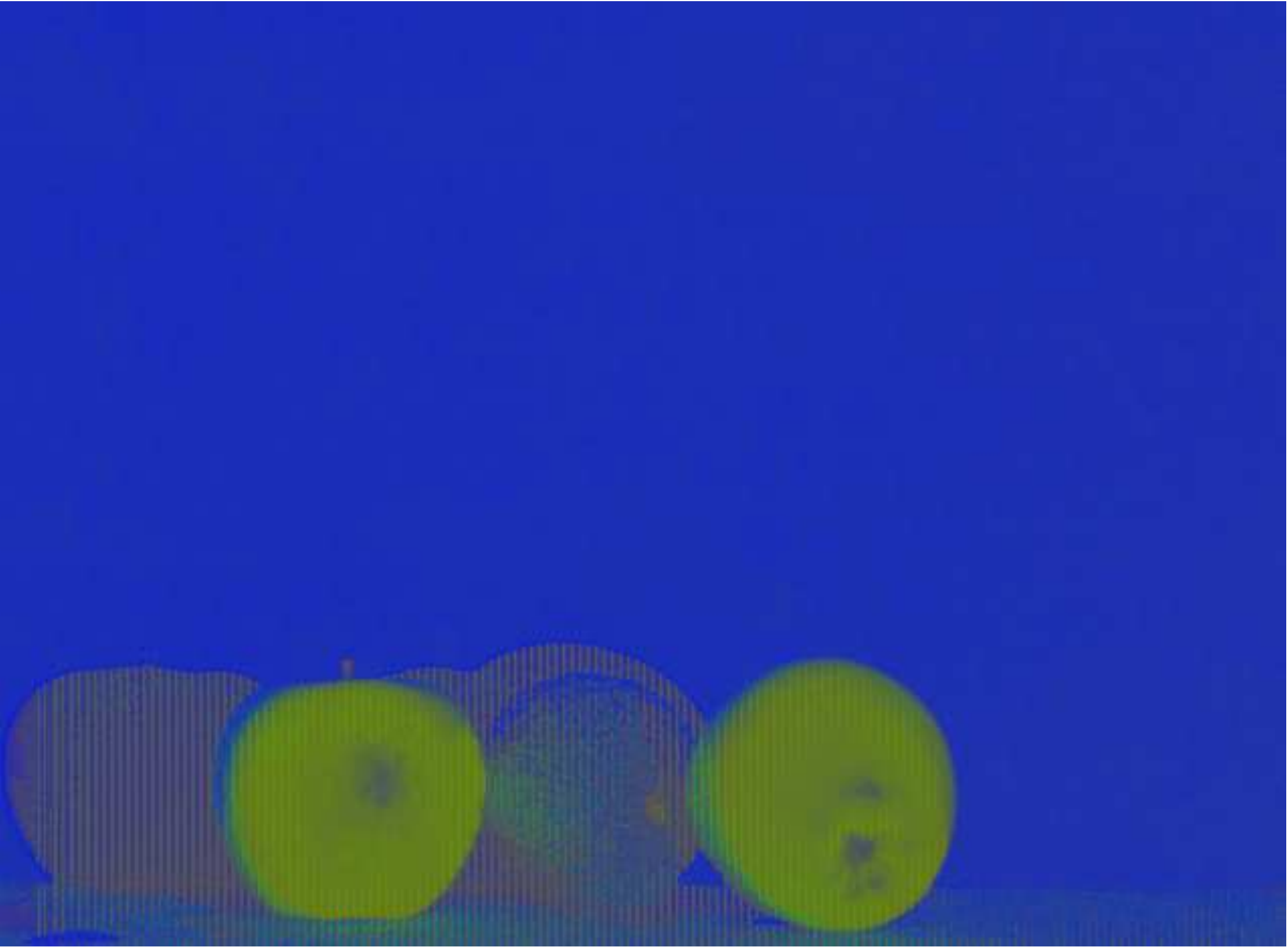}\ 
			\includegraphics[width=3.75cm]{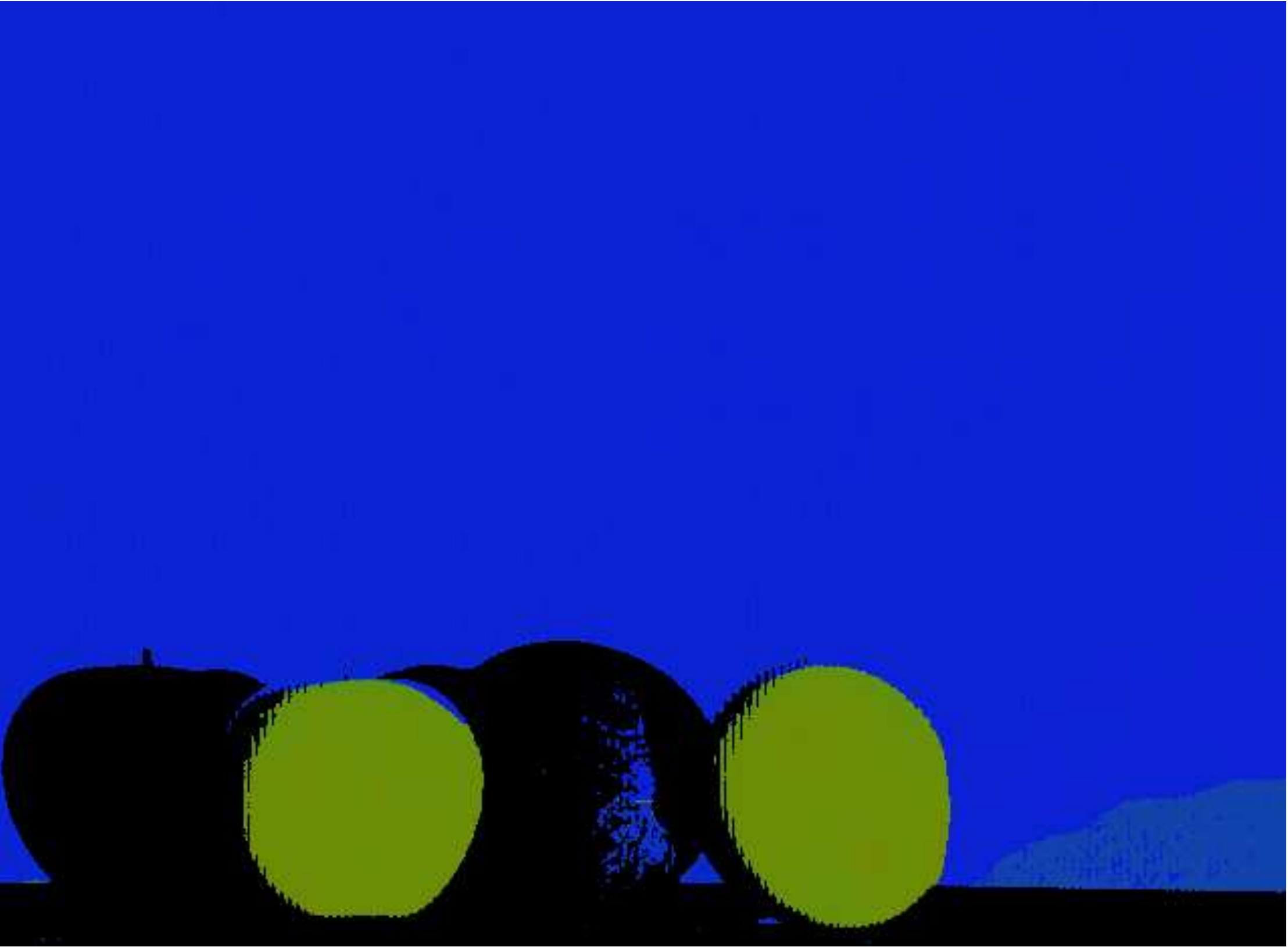}\ 
			
			\\
			
			\includegraphics[width=3.75cm]{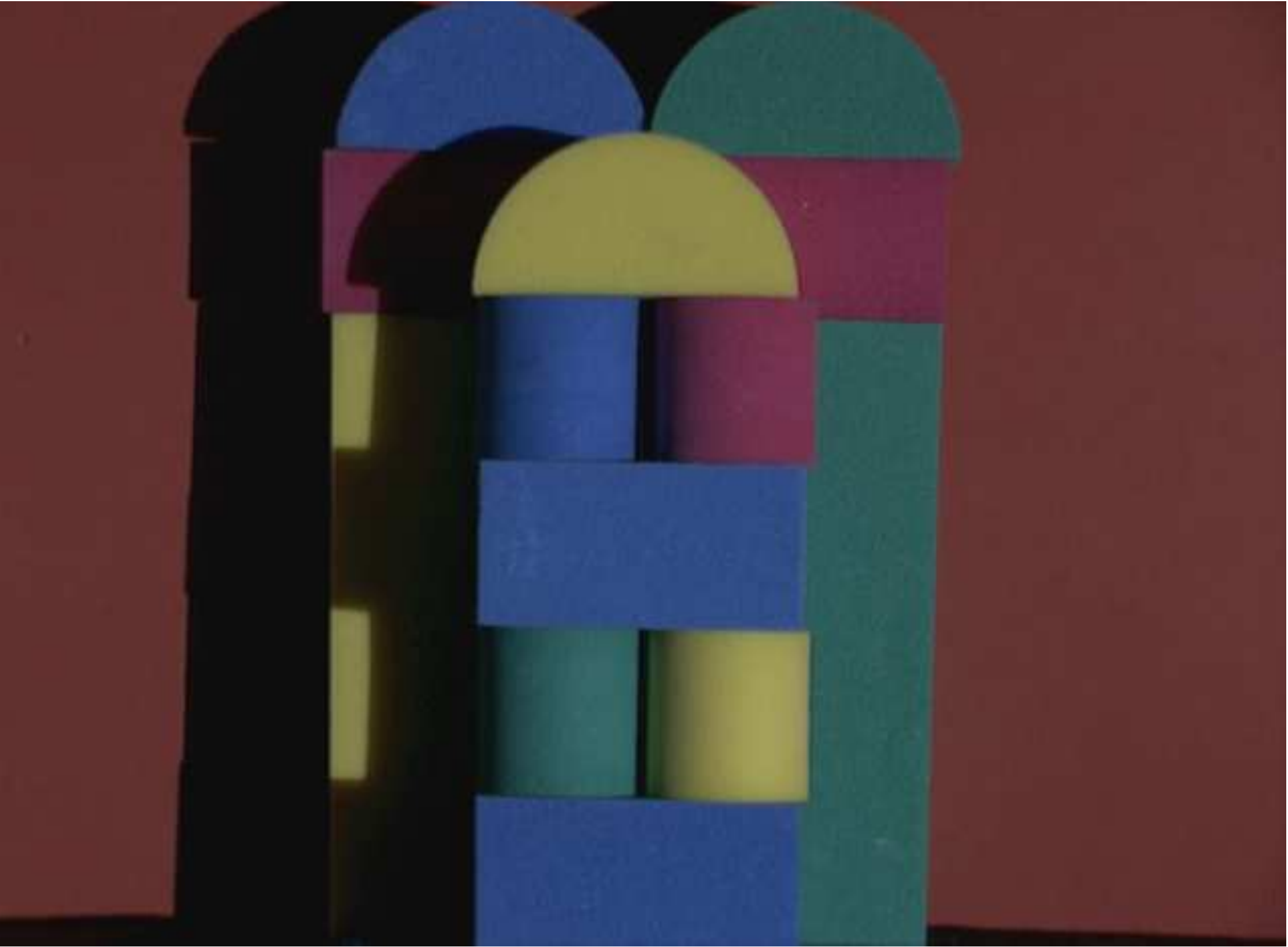}\ 
			\includegraphics[width=4cm]{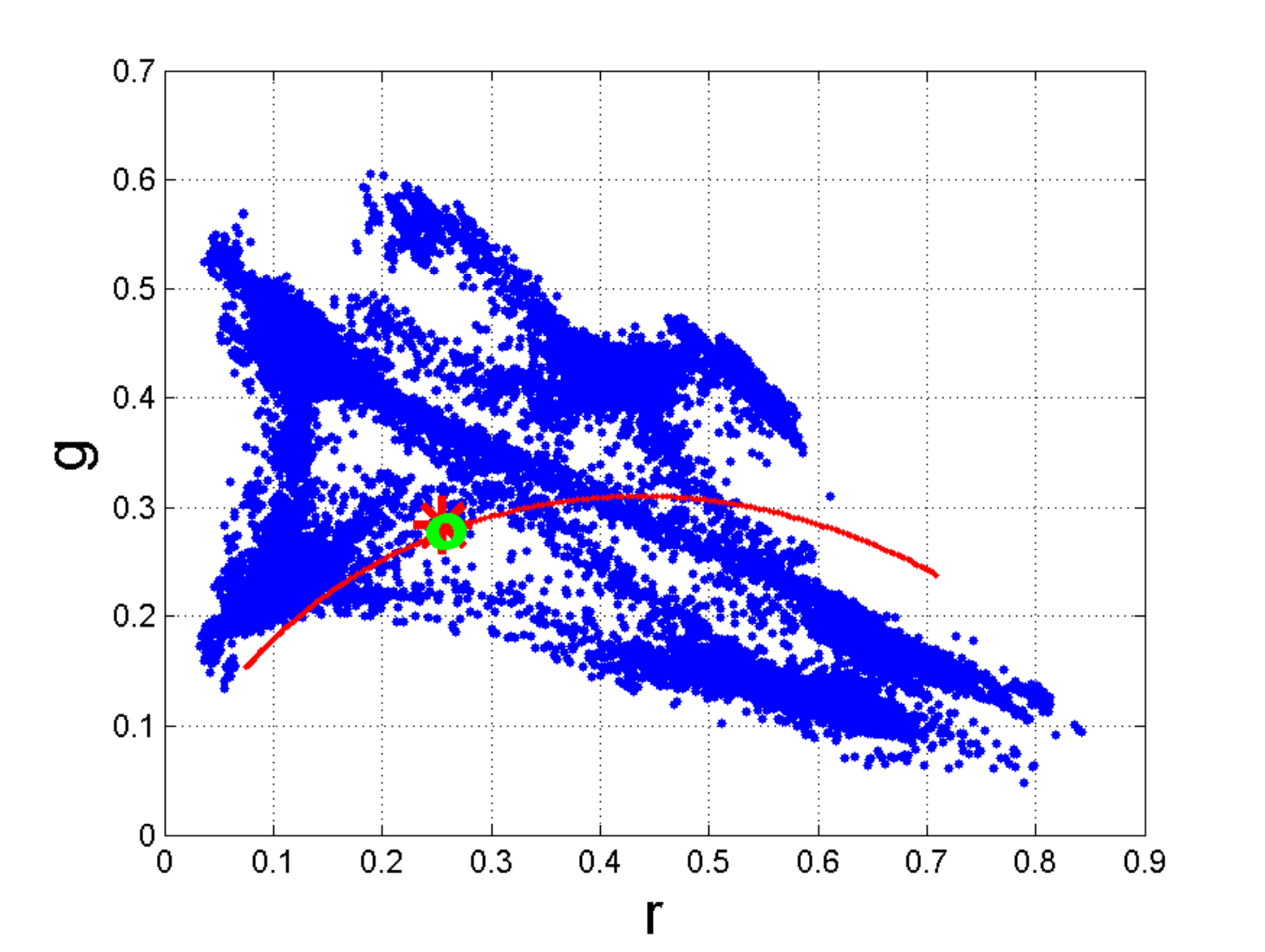}\ 
			\includegraphics[width=3.75cm]{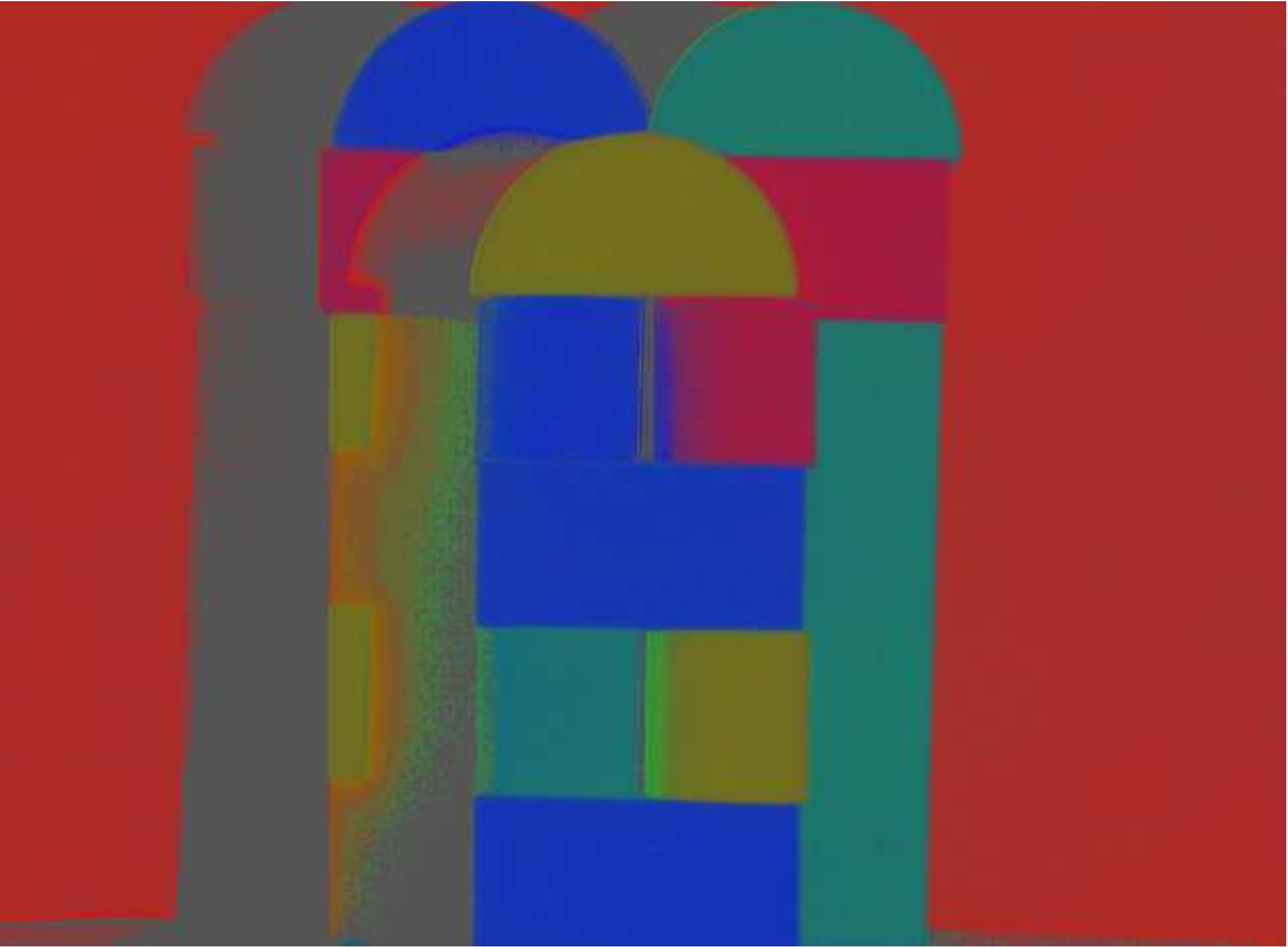}\ 
			\includegraphics[width=3.75cm]{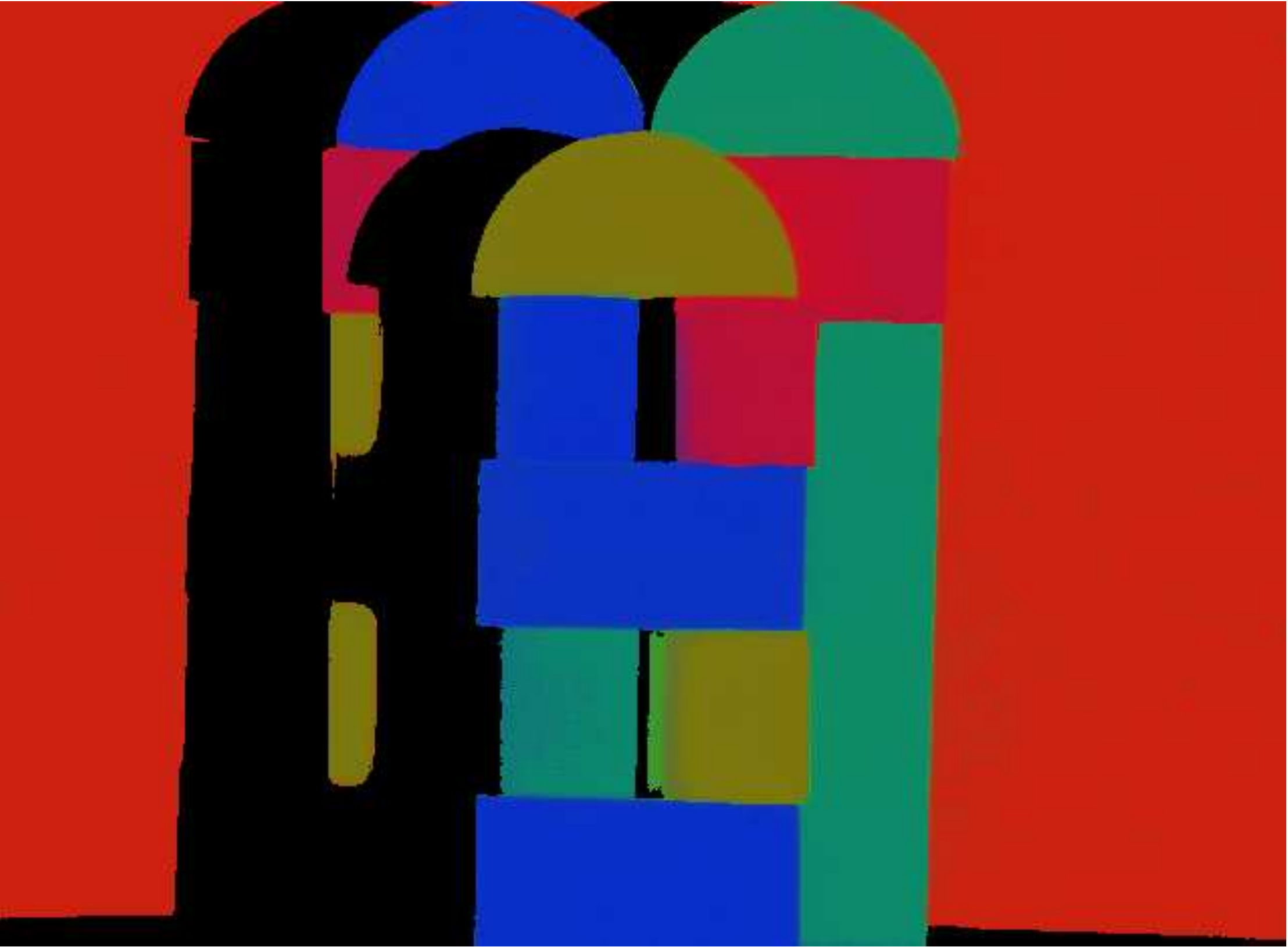}\ 
			
			\\
			
			\includegraphics[width=3.75cm]{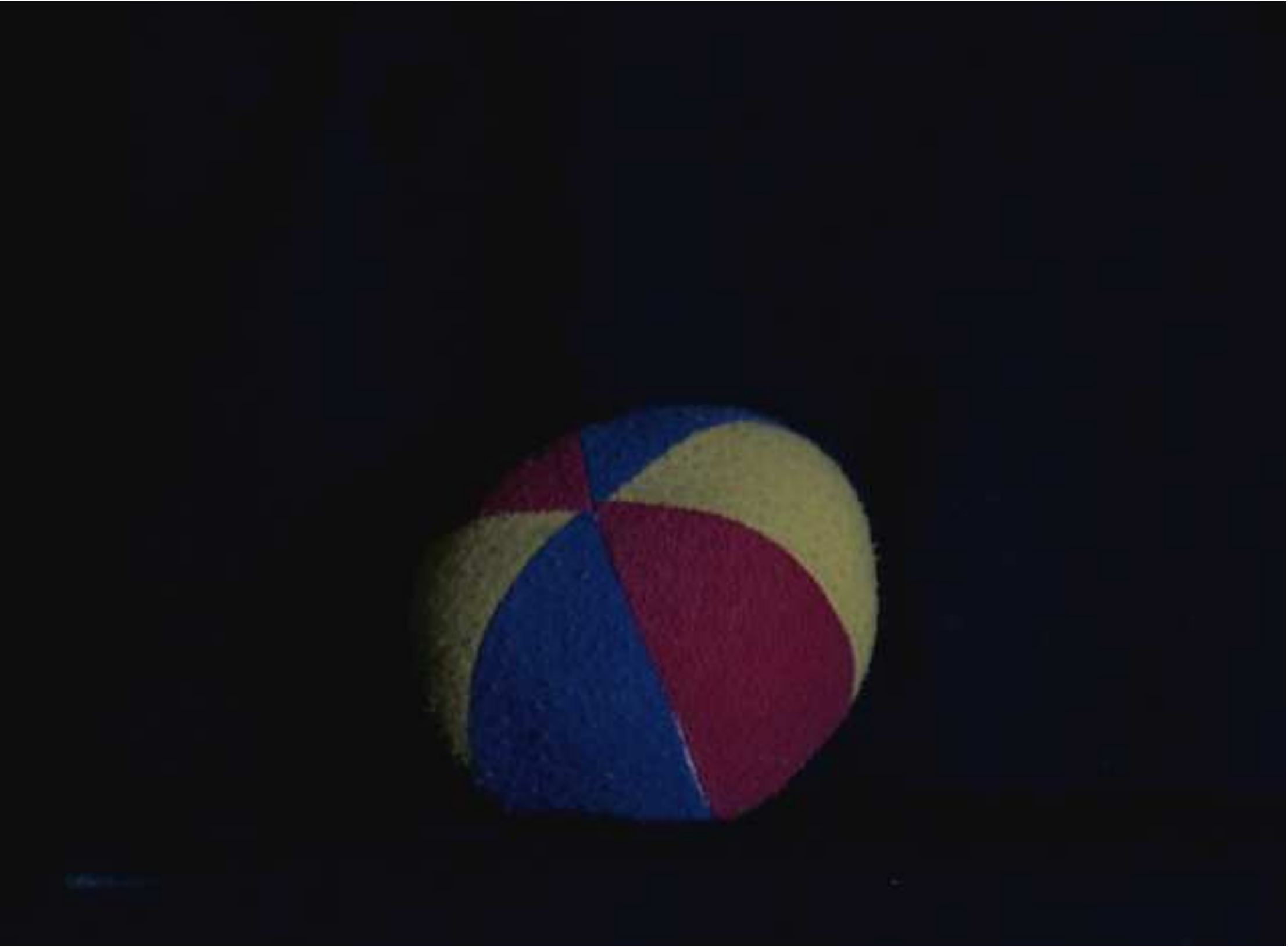}\ 
			\includegraphics[width=4cm]{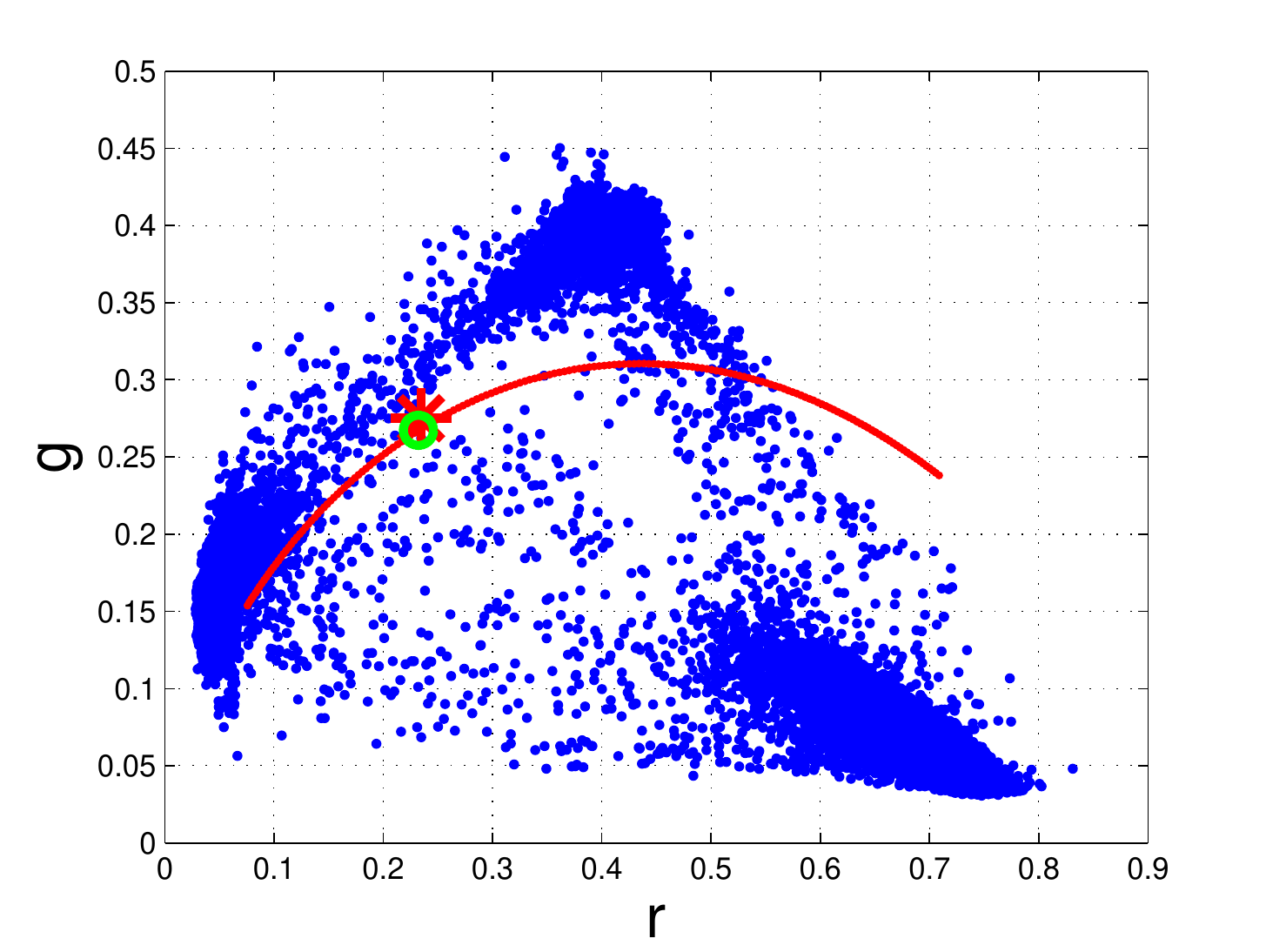}\ 
			\includegraphics[width=3.75cm]{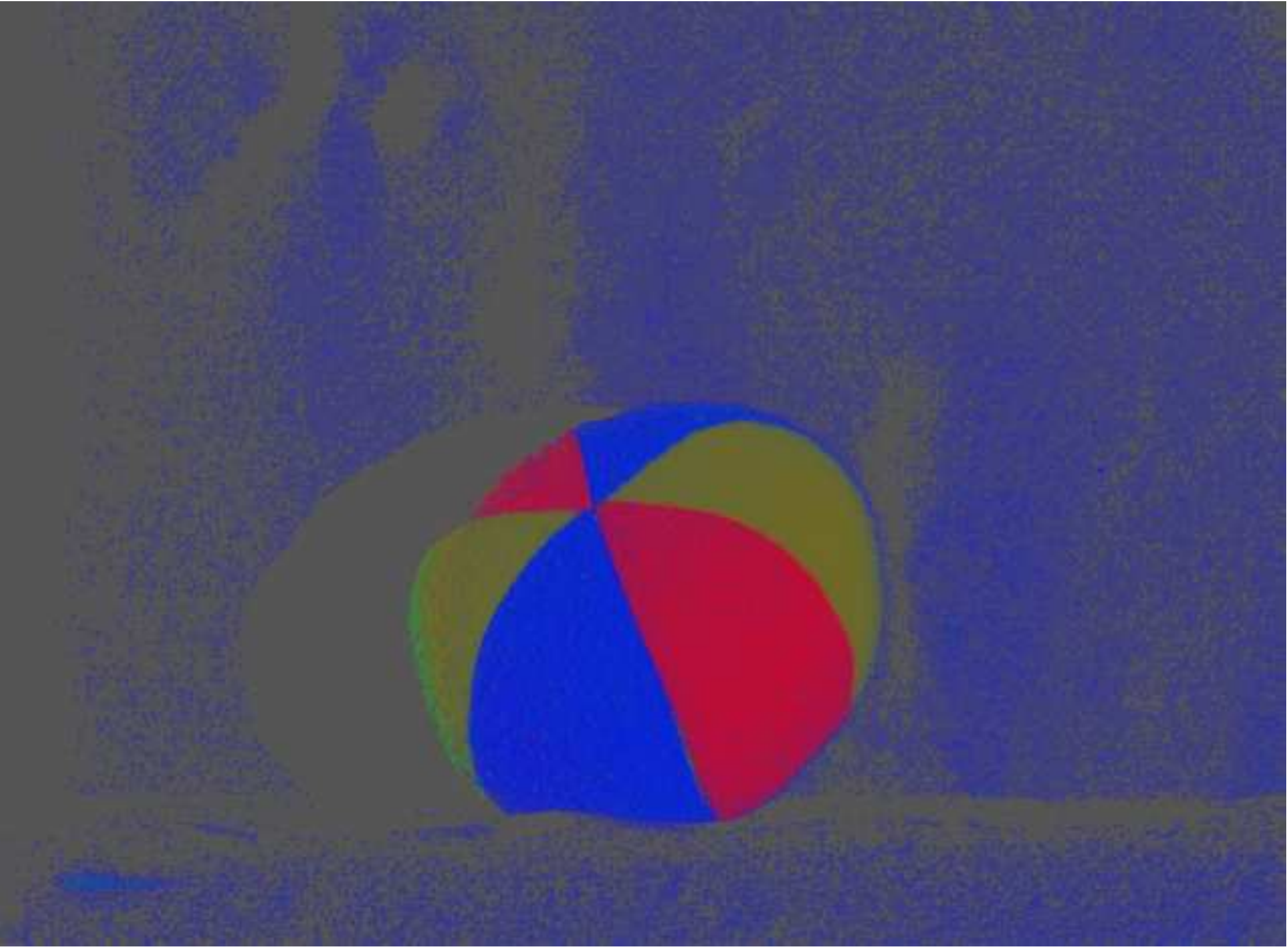}\ 
			\includegraphics[width=3.75cm]{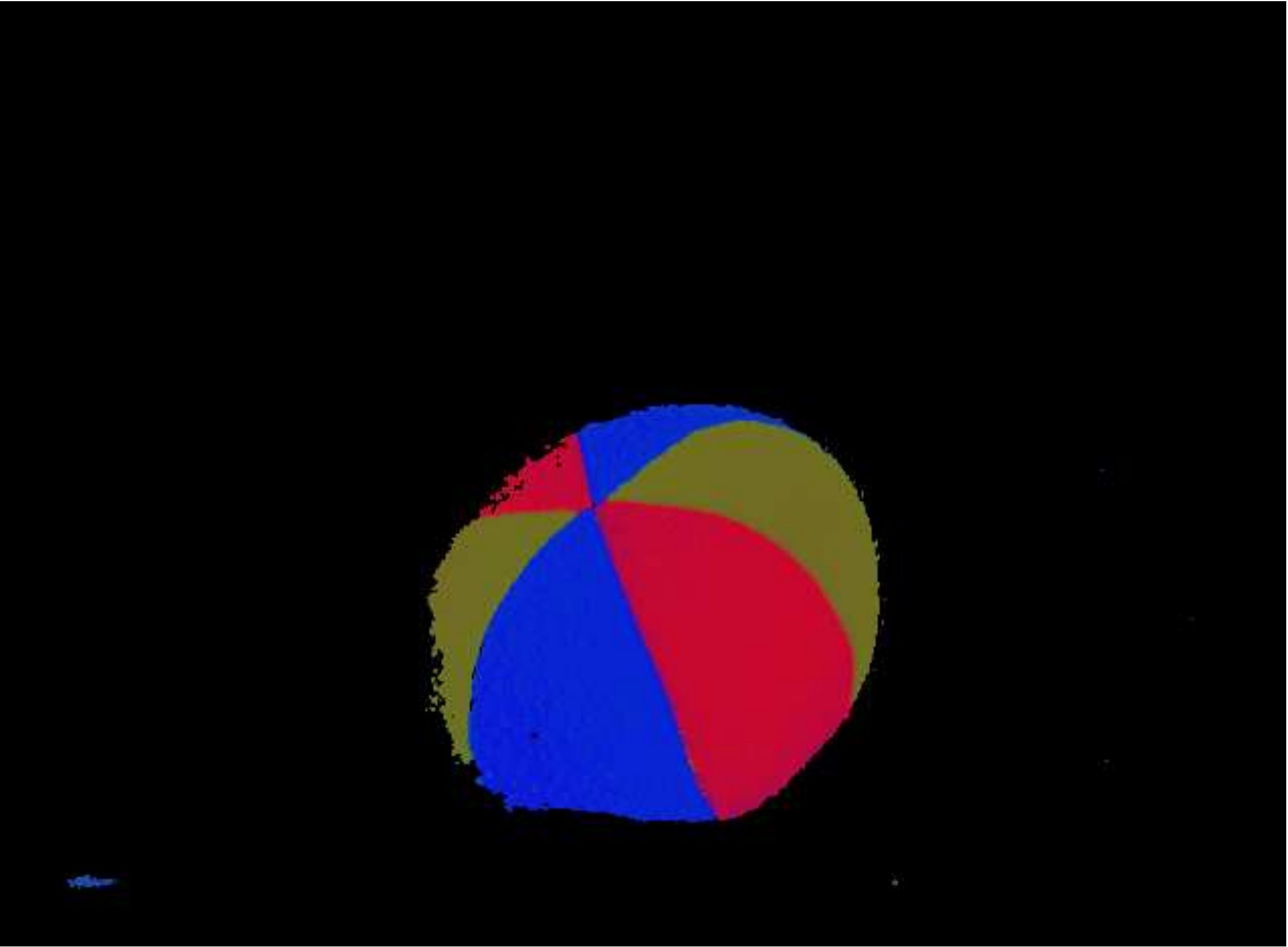}\

		\end{tabular}
		\vspace*{-0.25in}
		\caption{
			Left column: Input image; Second column: chromaticity and illuminant estimate;
			Third column: input image chromaticity shows
			highlights; Fourth column: proposed method removes shading and highlights.
		}
		\label{FIG:RESULTSPC}
	\end{center}
	\BUT
\end{figure}

Fig.~\ref{FIG:RESULTSPC} shows results, including finding the specular point and
generating a matte colour image, for 4 of input images: whereas the
original images' chromaticity clearly shows highlight effects and some shading,
output for the proposed method effectively eliminates these effects.

\section{Conclusion \label{SEC:CONCLUSIONCH6}}

In this paper we present a new camera calibration method aimed at
recovering parameters for the locus followed by illuminants in a special 2-D
chromaticity space.
The objective is to discover the colour-temperature of the illuminant in the
scene, for a new image not in the training set but captured using the calibrated
camera.

As a testing method to verify the validity of the proposed locus idea,
we compare illuminant recovery making use of the suggested locus as opposed to
not using it.  We determined that adding the locus constraint does indeed help
identify the scene illuminant. While the effect is not large, nonetheless the
experiments do provide a justification of the locus approach --- a new insight
in physics-based vision.

As an additional capability, we can subsequently generate a new version of the
input
image, shown as it would appear re-lit under new lighting conditions
by considering different illuminant values as the illuminant moves along the
specular-point locus.

In future work we will investigate how to make the method more robust to
illuminants that differ more substantially from Planckians.

\small
\bibliographystyle{osajnl}
\bibliography{thes-both}

\end{document}